\documentclass{article} %
\usepackage{iclr2023_conference}

\usepackage{amsmath,amsfonts,bm, xspace}
\usepackage{mathtools}

\def\textencoder{f_{txt}}
\def\imageencoder{f_{img}}
\def\index{\mathcal{H}}
\def\database{\mathcal{I}}
\def\dist{\mathbf{s}}
\def\image{\textnormal{I}}

\def\knn{kNN\xspace}
\def\noknn{\textit{no-kNN\,\xspace}}

\newcommand\knndiffusion{\textit{kNN-Diffusion}\xspace}
\newcommand\clip{CLIP\xspace}
\newcommand{\kld}[2]{\ensuremath{D_{KL}\infdivx{#1}{#2}}\xspace}
\DeclarePairedDelimiterX{\infdivx}[2]{(}{)}{%
  #1\;\delimsize|\delimsize|\;#2%
}

\DeclareMathOperator{\argmin}{arg\,min}

\def\eqref#1{equation~\ref{#1}}

\def\1{\bm{1}}

\DeclareMathAlphabet{\mathsfit}{\encodingdefault}{\sfdefault}{m}{sl}
\SetMathAlphabet{\mathsfit}{bold}{\encodingdefault}{\sfdefault}{bx}{n}

\usepackage{xr-hyper}
\usepackage{times}  %
\usepackage{helvet}  %
\usepackage{courier}  %
\usepackage{graphicx} %
\usepackage{natbib}  %
\usepackage{caption} %
\usepackage{algorithm}
\usepackage{algorithmic}

\usepackage[T1]{fontenc}
\usepackage[font=small,labelfont=bf,tableposition=top]{caption}

\DeclareCaptionLabelFormat{andtable}{#1~#2  \&  \tablename~\thetable}

\usepackage{amsmath}
\usepackage{amssymb}
\usepackage{comment}
\usepackage{multirow,bigdelim}
\usepackage{lipsum}
\usepackage{array}
\usepackage{wrapfig}

\usepackage{adjustbox}
\usepackage[percent]{overpic}
\usepackage{makecell}
\usepackage[normalem]{ulem}
\usepackage{blindtext}
\usepackage{xcolor}
\usepackage{soul}
\usepackage{subfig}
\usepackage{booktabs}
\newcommand\samethanks[1][\value{footnote}]{\footnotemark[#1]}

\usepackage[breaklinks=true,bookmarks=false,hyperfootnotes=false]{hyperref}
\usepackage{comment}
\usepackage{multirow,bigdelim}
\usepackage{lipsum}
\usepackage{array}
\usepackage{wrapfig}

\usepackage{adjustbox}
\usepackage[percent]{overpic}
\usepackage{makecell}
\usepackage[normalem]{ulem}
\usepackage{blindtext}
\usepackage{xcolor}
\usepackage{soul}
\usepackage{subfig}
\usepackage{booktabs}
\usepackage{amsmath}

\newcolumntype{C}[1]{>{\centering\let\newline\\\arraybackslash\hspace{0pt}}m{#1}}

\newcommand\imsize{2.2cm}

\newif\ifdraft
\drafttrue
\ifdraft
\newcommand{\oa}[1]{{\color{red}[\textbf{OA:} #1]}}
\newcommand{\sh}[1]{{\color{purple}[\textbf{SS:} #1]}}
\newcommand{\ap}[1]{{\color{blue}[\textbf{AP:} #1]}}

\newcommand{\yt}[1]{{\color{red}[\textbf{YT} #1]}}
\newcommand{\de}[1]{{\color{orange}[\textbf{DP:} #1]}}
\newcommand{\abc}[1]{{\color{green}[\textbf{AB:} #1]}}

\newcommand{\drop}[1]{}

\else
\newcommand{\oa}[1]{}
\newcommand{\sh}[1]{}
\newcommand{\ap}[1]{}
\newcommand{\yt}[1]{}
\newcommand{\de}[1]{}
\newcommand{\abc}[1]{}

\fi

\makeatletter
\DeclareRobustCommand\onedot{\futurelet\@let@token\@onedot}
\def\@onedot{\ifx\@let@token.\else.\null\fi\xspace}

\makeatother

\usepackage[bottom]{footmisc}

\usepackage[bottom]{footmisc}
\title{KNN-Diffusion: Image Generation via Large-Scale Retrieval}

\author{Shelly Sheynin\thanks{Equal Contribution} \quad Oron Ashual\samethanks \\ \quad Adam Polyak \quad Uriel Singer \quad Oran Gafni \quad Eliya Nachmani \quad Yaniv Taigman \\
Meta AI \\
{\tt\small \{shellysheynin,oron\}@meta.com}
}

\usepackage[capitalize, sort]{cleveref}
\crefname{section}{Sec.}{Secs.}
\Crefname{section}{Section}{Sections}
\Crefname{table}{Table}{Tables}
\crefname{table}{Tab.}{Tabs.}

\begin{document}

\maketitle
\begin{figure}[ht!]
\centering
\includegraphics[width=1.\linewidth]{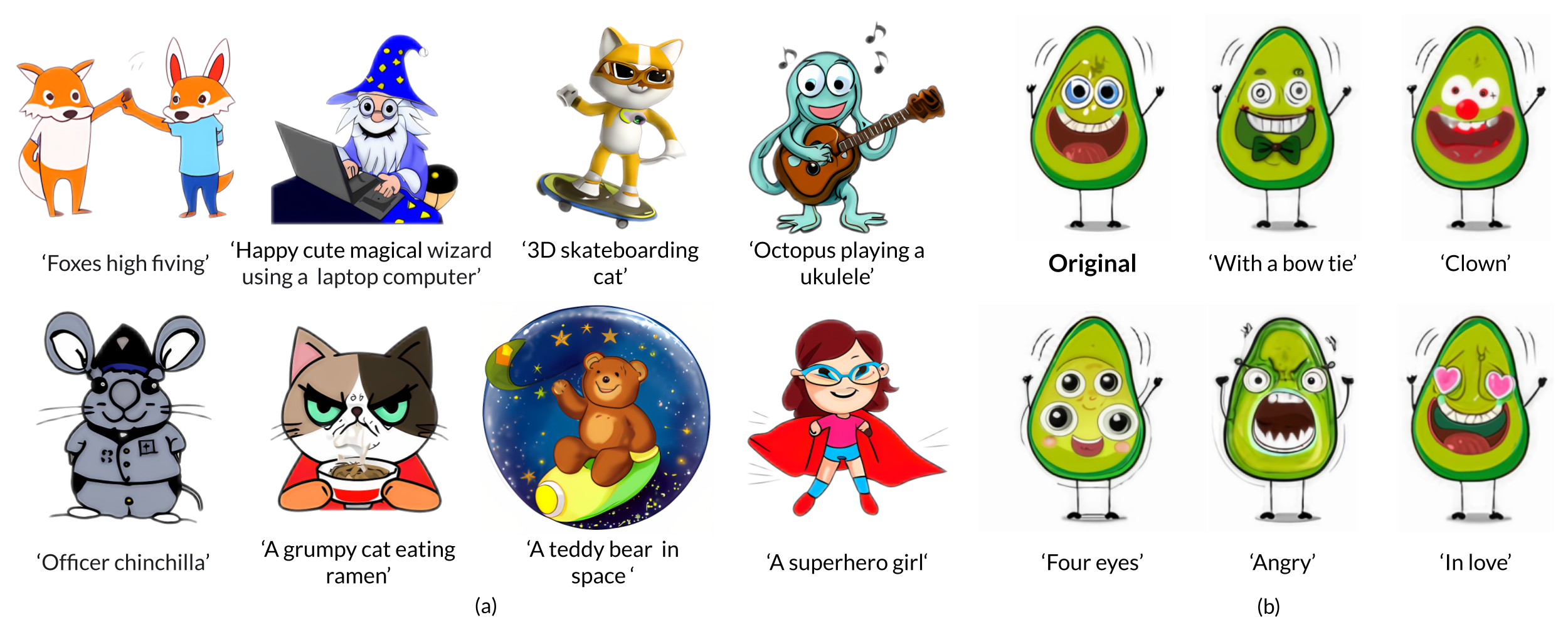}
\caption{(a) Samples of stickers generated from text inputs, (b) Semantic text-guided manipulations applied to the "Original" image without using edit masks. In both cases, our model was trained without any text data.} 
\label{fig:alpha_stickers_main}
\end{figure}
\vspace{.3cm}
\begin{abstract}
\label{section:abstract}
Recent text-to-image models have achieved impressive results. However, since they require large-scale datasets of text-image pairs, it is impractical to train them on new domains where data is scarce or not labeled.
In this work, we propose using large-scale retrieval methods, in particular, efficient $k$-Nearest-Neighbors (\knn), which offers novel capabilities: (1) training a substantially small and efficient text-to-image diffusion model without any text, (2) generating out-of-distribution images by simply swapping the retrieval database at inference time, and (3) performing text-driven local semantic manipulations while preserving object identity. To demonstrate the robustness of our method, we apply our \knn approach on two state-of-the-art diffusion backbones, and show results on several different datasets. As evaluated by human studies and automatic metrics, our method achieves state-of-the-art results compared to existing approaches that train text-to-image generation models using images only (without paired text data).

\end{abstract}

\section{Introduction}
\label{section:intoduction}
Large-scale generative models have been applied successfully to image generation tasks~\citep{gafni2022make, dalle, glide, saharia2022photorealistic, yu2022scaling}, and have shown outstanding capabilities in extending human creativity using editing and user control. However, these models face several significant challenges: (i) \textbf{Large-scale paired data requirement}. To achieve high-quality results, text-to-image models %
rely heavily on large-scale datasets of (text, image) pairs collected from the internet. %
Due to the requirement of paired data, these models cannot be applied to new or customized domains with only unannotated images.
(ii) \textbf{Computational cost and efficiency}. Training these models on highly complex distributions of natural images usually requires scaling the size of the model, data, batch-size, and training time, which makes them challenging to train and less accessible to the community. %
Recently, several works proposed text-to-image models trained without paired text-image datasets. \citet{fusedream} performed a direct optimization to a pre-trained model based on a \clip loss~\citep{clip}. Such approaches are time-consuming, since they require optimization for each input. \citet{lafite} proposed training with \clip image embedding perturbed with Gaussian noise. However, to achieve high-quality results, an additional model needs to be trained with an annotated text-image pairs dataset.

\noindent In this work, we introduce a novel generative model, \knndiffusion, which tackles these issues and progresses towards more accessible models for the research community and other users. Our model leverages a large-scale retrieval method, $k$-Nearest-Neighbors (\knn) search, in order to train the model without any text data.
Specifically, our diffusion model is conditioned on two inputs: (1) image embedding (at training time) or text embedding (at inference), extracted using the multi-modal \clip encoder, and (2) \knn embeddings, representing the $k$ most similar images in the \clip latent space.
During training, we assume that no paired text is available, hence condition only on \clip image embedding and on $k$ additional image embeddings, selected using the retrieval model. At inference, only text inputs are given, so instead of image embeddings, we use the text embedding that shares a joint embedding space with the image embeddings. Here, the \knn image embeddings are retrieved using the \textit{text} embeddings.

\begin{figure*}[t!]
\centering
\includegraphics[width=1.\linewidth]{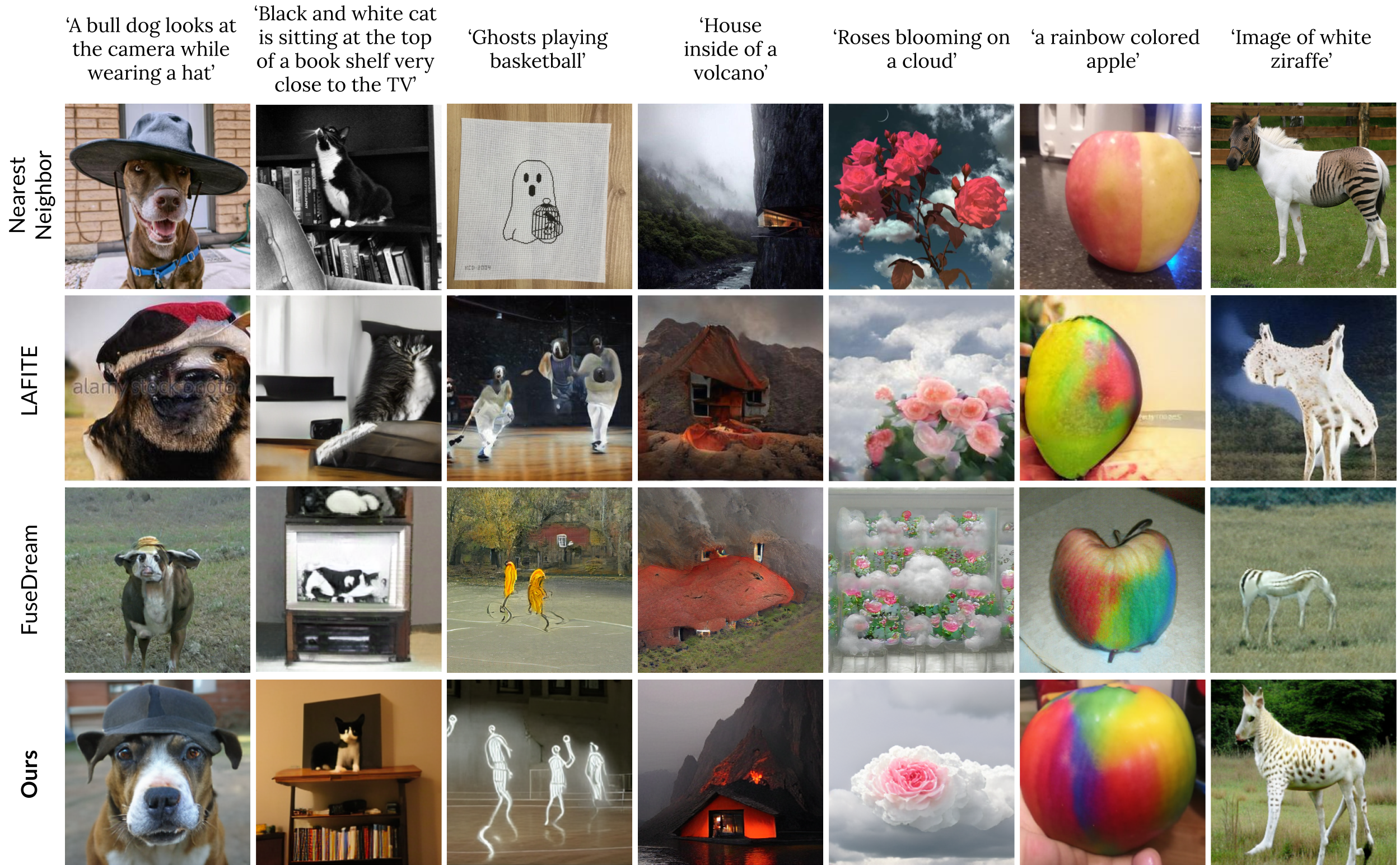} 
\caption{Qualitative comparisons with previous work. \textit{Nearest Neighbor} is the first nearest neighbor of the text-input in the PMD dataset.} 
\label{fig:cmd_hard}
\vspace{-0.5cm}
\end{figure*}

The additional \knn image embeddings have three main benefits: (1) they extend the distribution of conditioning embeddings and ensure the distribution is similar in train and inference, thus helping to bridge the gap between the image and text embedding distributions; (2) they teach the model to learn to generate images from a target distribution by using samples from that distribution. This capability allows generalizing to different distributions at test time and generating out-of-distribution samples; (3) they hold information that
does not need to be present in the model, which allows it to be substantially smaller. We demonstrate the effectiveness of our \knn approach in Sec.~\ref{section:experiments}. 

To assess the performance of our method, we train our model on two large-scale datasets: the Public Multimodal Dataset~\citep{singh2021flava} and an image-only stickers dataset collected from the Internet. We show state-of-the-art zero-shot capabilities on MS-COCO~\citep{lin2014microsoft}, LN-COCO~\citep{pont2020connecting} and CUB~\citep{wah2011caltech}.
To further demonstrate the advantage of using retrieval methods in text-to-image generation, we train two diffusion backbones using our \knn approach: continuous~\citep{ramesh2022hierarchical} and discrete~\citep{Gu2021VectorQD}. In both cases we outperform the model that is trained without \knn. In comparison to alternative methods presented in Sec.~\ref{section:experiments}, we achieve state-of-the-art results in both human evaluations and FID score, with 400 million parameters and 7 second inference time. 

\begin{figure}[t!]
\centering
\includegraphics[width=1.\linewidth]{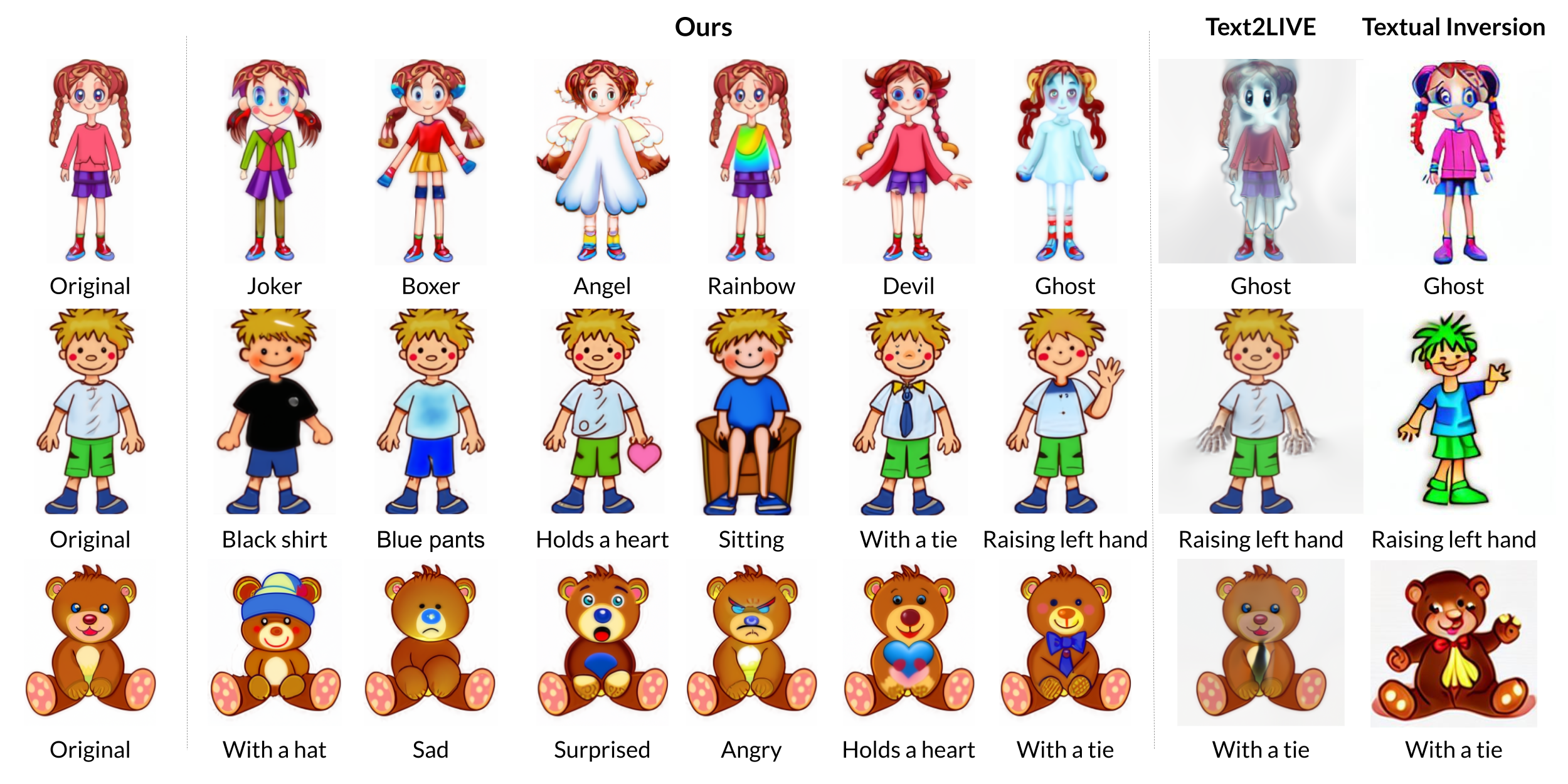}
\caption{Results for text-guided image manipulations without using masks. The original image is shown in the left column, our manipulated images are shown in the center. The images of ~\cite{bar2022text2live,gal2022image} were generated using the authors' official code. The full comparison is available in the supplement.}
\label{fig:human_manipulation}
\vspace{-.4cm}
\end{figure}

Lastly, we introduce a new approach for local and semantic manipulations that is based on \clip and \knn, without relying on user-provided masks. Specifically, we fine-tune our model to perform local and complex modifications that satisfies a given target text prompt. For example, given the teddy bear's image in Fig.~\ref{fig:human_manipulation}, and
the target text "holds a heart", our method automatically locates the local region that should be modified and synthesizes a high-resolution manipulated image in which (1) the teddy bear's identity is accurately preserved and (2) the manipulation is aligned with the target text. We demonstrate our qualitative advantage by comparing our results with two state-of-the-art models, Text2Live~\citep{bar2022text2live} and Textual Inversion~\citep{gal2022image}, that perform image manipulations without masks (Fig.~\ref{fig:human_manipulation}, ~\ref{fig:text2live_compare} and~\ref{fig:textual_compare}).

We summarize the contributions of this paper as follows: (1) We propose \knndiffusion, a novel and efficient model that utilizes a large-scale retrieval method for training a text-to-image model without using text. 
(2) We demonstrate efficient out-of-distribution generation, which is achieved by substituting retrieval databases. 
(3) We present a new approach for local and semantic image manipulation, without utilizing masks.
(4) We evaluate our method on two diffusion backbones, discrete and continuous, as well as on several datasets, and present state-of-the-art results compared to baselines trained using image-only datasets.

\section{Related Work}
\smallskip  \noindent{\bf Text-to-image models.\quad}
Text-to-image generation is a well-studied task that focuses on generating images from text descriptions.
While GANs~\citep{xu2018attngan,zhu2019dm,zhang2021cross} and VQ-VAE Transformer-based methods~\citep{dalle,gafni2022make,yu2022scaling,cogview} have shown remarkable results, recently impressive results have been attained with discrete~\citep{Gu2021VectorQD} and continuous~\citep{glide,saharia2022photorealistic,ramesh2022hierarchical,rombach2022high} diffusion models. Most recent works trained diffusion models conditioned on text embeddings extracted using a pre-trained text encoder~\citep{saharia2022photorealistic,yu2022scaling} or image embedding extracted using \clip ~\citep{ramesh2022hierarchical}. 
While producing impressive results, all previous works described above are supervised and trained with paired text-image datasets. 

Several works have proposed training text-to-image models without text data. FuseDream~\citep{fusedream} proposed a direct optimization to a pre-trained generative model based on \clip loss. This method relies on a pre-trained GAN and requires a time-consuming optimization process for each image.
LAFITE~\citep{lafite} recently demonstrated text-to-image generation results without requiring paired text-image datasets. Here, the embeddings of \clip are used interchangeably at train and test time to condition a GAN-based model. The joint text-image embedding enables inference given a text input, whereas in training the model is fed with the visual embedding only. However, the gap between the text and image distributions in the joint embeddings space leads to results with substantially lower quality, as we show in our experiments. To overcome this gap, LAFITE added noise to the image embeddings during training. Our remedy to this gap in distributions is to condition the model on the retrieval of an actual image embeddings, using a text-image joint space.%

\smallskip \noindent{\bf Retrieval for generation.\quad} The Information Retrieval (IR) literature tackles the challenge of retrieving a small amount of information from a large database, given a user's query. %
While it has been shown that adding more training data improves performance, this comes at a cost, as it requires collecting and training on more data. 
A simple, yet efficient retrieval mechanism is to retrieve the $K$ nearest neighbors (\knn) between the query and the entities in the database in some pre-calculated embedding space~\citep{bijalwan2014knn}.
The database allows the model to leverage extensive world-knowledge for its specific task \cite{borgeaud2021improving}. Recently, language models were augmented with a memory component, allowing them to store representations of past inputs~\citep{wu2022memorizing}. The latter were then queried using a lookup operation, improving performance in various benchmarks and tasks.
Retrieval models have been used for various tasks in learning problems, for example, language modeling~\citep{borgeaud2021improving}, machine translation~\citep{gu2018search}, question answering~\citep{lee2019latent} and image generation~\citep{tseng2020retrievegan}. RetrieveGAN~\citep{tseng2020retrievegan} uses a differentiable retrieval module for image generation from a scene description, RetrievalFuse~\citep{siddiqui2021retrievalfuse} proposed a neural 3D scene reconstruction based on a retrieval system. %
In this work we utilize the \knn retrieval mechanism over the shared text-image embedding space, CLIP~\citep{clip}. Using extensive ablation studies, we show the importance of the retrieval model both for training and inference, and demonstrate its large impact on performance.

\begin{figure}[!t]
\begin{minipage}[t!]{\textwidth}
\begin{minipage}[b]{0.44\textwidth}
\centering
\includegraphics[width=1.0\linewidth]{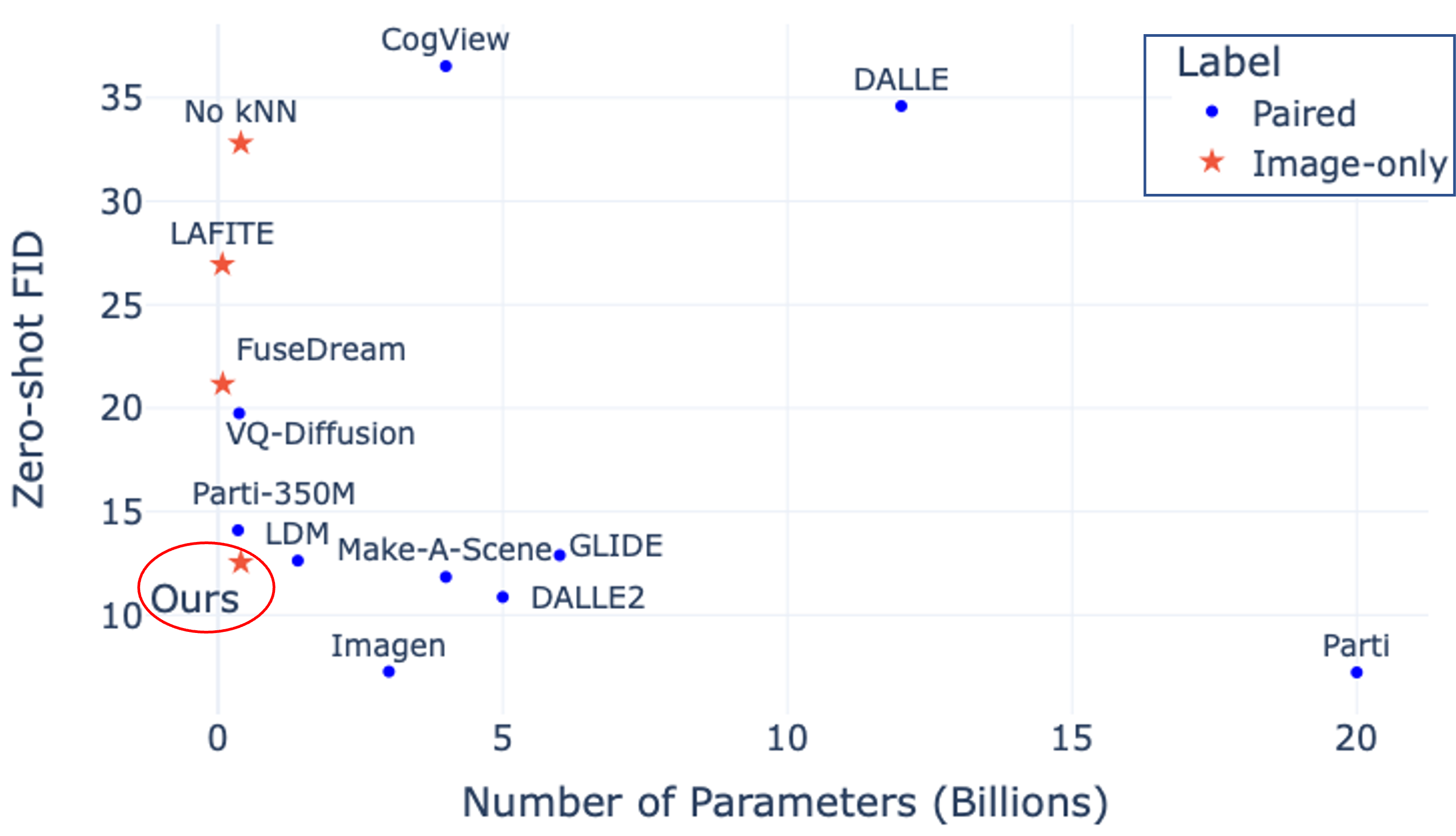} 
\captionof{figure}{FID comparison on MS-COCO, including models trained on image-only datasets and text-image datasets.}
\label{all_models_fid_coco}
\end{minipage}
\hfill
\quad
\begin{minipage}[b]{0.54\textwidth}
\centering
  \captionof{table}{Results for zero-shot Text-to-Image generation on the MS-COCO, CUB and LN-COCO test sets. Image-quality and Text-alignment report the percentage of majority votes in favor of our method when comparing between a certain model and ours.}
\scalebox{0.85}{
\setlength{\tabcolsep}{1.0pt}
\begin{tabular}[b]{l|ccc|ccc|ccc}
    \toprule
     & \multicolumn{3}{c}{MS-COCO} &  \multicolumn{3}{|c}{CUB} & \multicolumn{3}{|c}{LN-COCO} \\
    \multicolumn{1}{c|}{Model} & \multicolumn{1}{c}{FID$\downarrow$} & \begin{tabular}{@{}c@{}}Im.\\qual.\end{tabular} & \begin{tabular}{@{}c@{}}Txt\\align.\end{tabular} & 
    \multicolumn{1}{c}{FID$\downarrow$} &
    \begin{tabular}{@{}c@{}}Im.\\qual.\end{tabular} & \begin{tabular}{@{}c@{}}Txt\\align.\end{tabular} & 
    \multicolumn{1}{c}{FID$\downarrow$} &
    \begin{tabular}{@{}c@{}}Im.\\qual.\end{tabular} & \begin{tabular}{@{}c@{}}Txt\\align.\end{tabular} \\
    \noalign{\smallskip}
    \midrule
    \small{LAFITE}  & $26.9$ & $72.1$ & $65.3$ & $89.7$ & $74.0$ & $59.6$ & $42.8$ & $68.4$  & $61.9$ \\
     \small{FuseDream} &  $21.2$ &$64.0$ & $79.3$ & $50.2$ & $79.1$ & $60.9$ & $37.5$ & $71.1$ & $59.0$ \\
     \small{\noknn} & $32.8$ & $70.8$ & $68.3$ & $95.1$ & $81.0$ & $61.2$ & $65.0$ & $61.4$ & $59.8$ \\
     \small{Ours} & $\textbf{12.5}$ & - &-  & $\textbf{42.9}$ & - & - & $\textbf{35.6}$ & - & -\\
\bottomrule
\end{tabular}
}
\label{tab:zero_shot_coco}
\end{minipage}
\end{minipage}
\vspace{-0.4cm}
\end{figure}
\setlength{\tabcolsep}{1.4pt}

\smallskip \noindent{\bf Multi-modal feature learning.\quad} %
Learning a joint and aligned feature space for several modalities is challenging, as it requires alignment between the modalities (paired datasets), whose distributions may vary. Specifically, the joint feature space of vision-and-language has been a long-standing problem in artificial intelligence. CLIP~\citep{clip} is a method that successfully tackled this challenge, by leveraging contrastive learning over a large dataset of aligned text-image pairs. 
Several works, such as BLIP~\citep{BLIP},~\citep{SLIP} and FLAVA~\citep{singh2021flava}, followed this idea and further improved the joint representation. The joint representation was shown to hold a strong semantic alignment between the two modalities, enabling image generation~\citep{fusedream, wang2022clip}, image manipulation~\citep{patashnik2021styleclip, avrahami2022blended}, and image captioning~\citep{mokady2021clipcap}.
In this work we leverage the joint representation in two ways: (i) enabling textless training with only visual data, while using text at inference time, and (ii) creating an efficient embedding space for the use of the retrieval model.

\begin{figure*}[t!]
\centering
\includegraphics[width=1.0\linewidth]{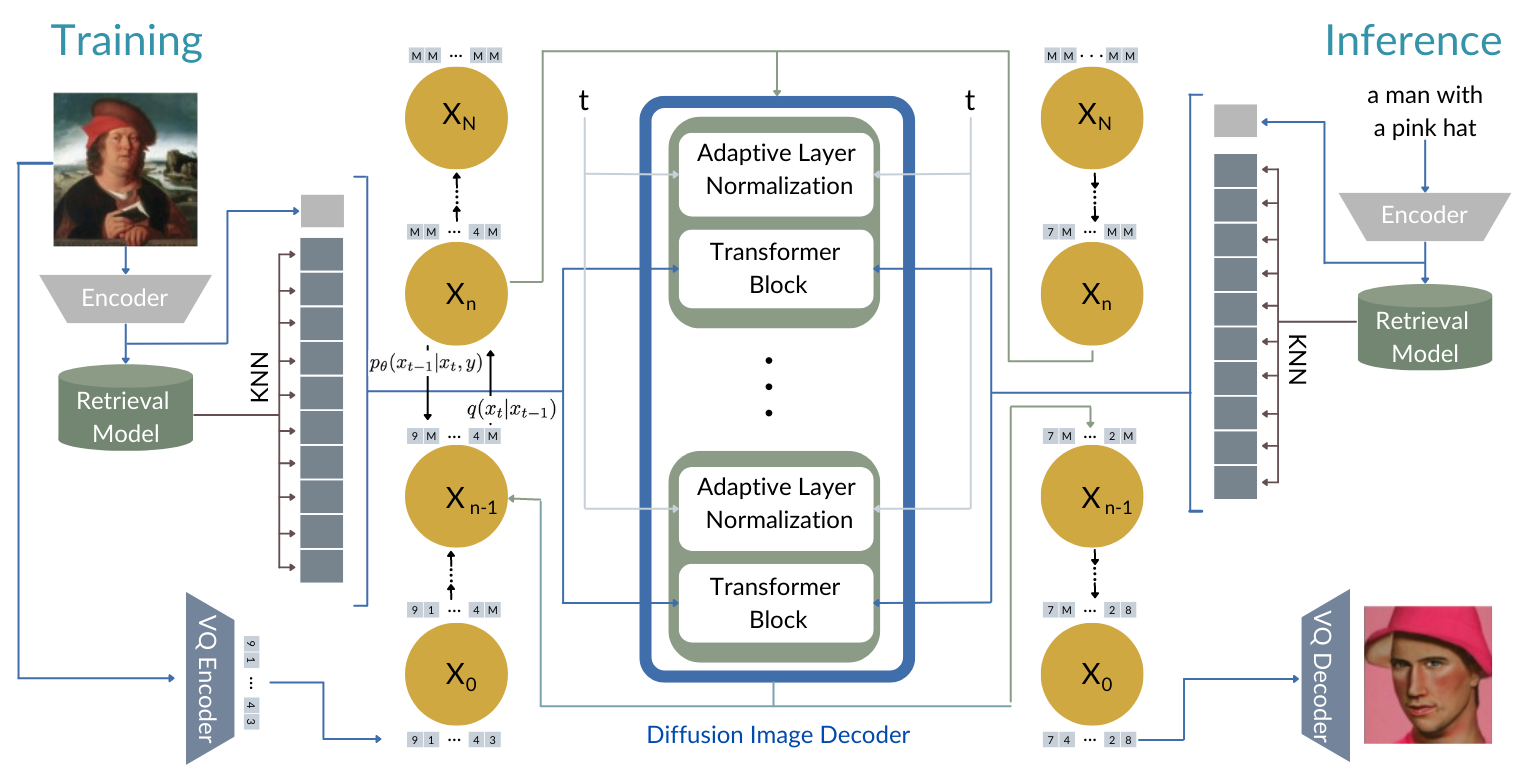}
\caption{\textbf{The overall framework of Discrete \knndiffusion.} During training, a diffusion-based transformer backbone is conditioned on $\imageencoder(\image)$ and its K nearest neighbors in index $\index$. %
At inference time, given an input text $\textbf{t}$, the condition to the process is similarly given by the K nearest neighbors, but of $f_{txt}(\textbf{t})$.}
\label{fig:architecture}
\vspace{-5.5pt}
\end{figure*}

\section{Method}
Our main goal is to facilitate language-guided generation of user-specified concepts while using an images-only dataset (without paired text data) during training.
A possible way to achieve this goal is to use a shared text-image encoder that will map text-image pairs into the same latent space, thus allowing training with an image embedding, and inferring from text embedding. A candidate for this encoder is \clip, which has been trained with a contrastive loss on a large-scale dataset of text-image pairs.
However, as we show quantitatively in Tab.~\ref{tab:zero_shot_coco},~\ref{tab:stickers} and qualitatively in Fig.~\ref{fig:cmd},~\ref{fig:stickers}, \clip embeddings alone cannot accurately bridge the gap between the text and image distributions. In order to reduce this gap, several methods have been proposed. The closest work to ours is LAFITE, which perturbs the \clip image embedding with adaptive Gaussian noise. %
As a result, the model is expected to be more robust to differences in the distributions. %
Under the assumption that there is a large paired text-image dataset, ~\citet{ramesh2022hierarchical} have proposed a prior that is used during inference, and is trained to generate possible \clip image embeddings from a given text caption. %

In this regard, we propose using a large-scale and non-trainable image embedding index as an integral part of the diffusion process. Our proposed method, \knndiffusion, assumes that only image data is provided during training. 
As shown in Fig.~\ref{fig:architecture}, our model is comprised of three main components: (1) A multi-modal text-image encoder (\clip); (2) A retrieval model - A data structure containing image embeddings, which is indexed for a fast \knn search; (3) An image generation network - A trainable diffusion-based image generation model, conditioned on the projected retrievals. 
For both training and inference, the image generation network is conditioned on $K$ additional image embeddings, chosen using the retrieval model to ensure a similar distribution of the condition in training and inference. The following sections describe these components, and additional details can be found in Sec.~\ref{subsection_supp:continuous} and~\ref{subsection_supp:discrete} in the supplement.

\smallskip \noindent{\bf Retrieval model.\quad}
Our retrieval model has three non-trainable modules: a pre-trained text encoder $\textencoder$, a pre-trained image encoder $\imageencoder$ and an index $\index$. The encoders map text descriptions and image samples to a joint multi-modal $d$-dimensional feature space $\mathbb{R}^{d}$. The index stores an efficient representation of the images database - $\index := \{\imageencoder(i)\in\mathbb{R}^{d} | i\in \database\}$ where $\database$ denotes the dataset of images. During training, we use the index to efficiently extract the $k$ nearest neighbors in the feature space of the image embedding $\imageencoder(\image)\in\mathbb{R}^{d}$ - $\text{knn}_{img}(\image, k):=\argmin^k_{h \in \index}\dist(\imageencoder(\image), h)$ where $\dist$ is a distance function and $\argmin^k$ output the minimal $k$ elements. The set $\{\imageencoder(\image), \text{knn}_{img}(\image, k)\}$ is used as the condition to the generative model. During inference, given a query text $t$, an embedding $\textencoder(t)$ is extracted. The generative model is conditioned on this embedding and its $k$ nearest neighbors from the database - $\text{knn}_{txt}(t, k):=\argmin^k_{h \in \index}\dist(\textencoder(t), h)$.

\begin{figure}[t!]
\centering
\subfloat[Training]
    {\includegraphics[page=1,width=3cm]{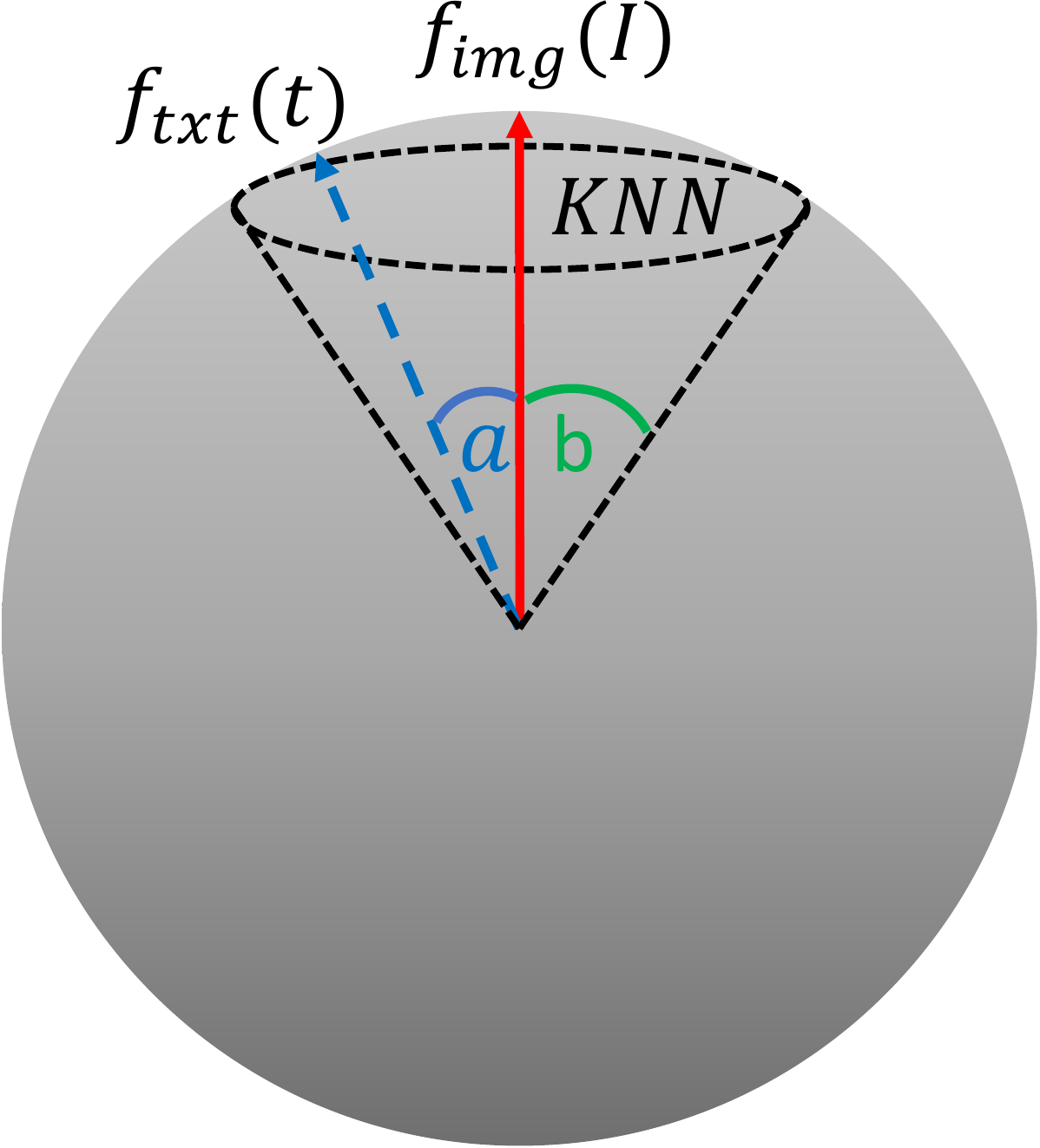}}
\hspace{20pt}
\subfloat[Inference]
    {\includegraphics[page=2,width=3cm]{resources/images/ball.pdf}}
\vspace{-.1cm}
\caption{During training, only the image $\image$ is given (red), whereas during inference only the text $t$ is given (blue). In order to bridge the gap between the two distributions during training, we leverage the K nearest neighbors that should have a large enough distribution (dashed cone) to cover the potential text embedding (i.e. $cos(b) < cos(a)$). During inference, the opposite is applied.}
\label{fig:ball}
\vspace{-.4cm}
\end{figure}

During training, we add embeddings of real images, by applying the retrieval method to the input image embedding. The extracted \knn should have a large enough distribution to cover the potential text embedding.
During inference, the \knn are retrieved using the text embedding (See Fig.~\ref{fig:ball}). In all of our experiments we use \clip as the encoders $\textencoder$ and $\imageencoder$, and the cosine similarity metric as the distance function $\dist$. Implementation details can be found in Sec.\ref{subsection_supp:retieval} of the supplement.

\smallskip \noindent{\bf Image generation network.\quad} In order to demonstrate the robustness of our method, we apply our \knn approach on two different diffusion backbones: Discrete ~\citep{Gu2021VectorQD} and Continuous ~\citep{glide,sohl2015deep,ho2020denoising,dhariwal2021diffusion}. Although very different in practice, these models share the same theoretical idea. Let $x_0\sim q(x_0)$ be a sample from our images distribution. A forward process of a diffusion model $q(x_n|x_{n-1})$ is a Markov chain that adds noise at each step. The reverse process, $p_\theta(x_{n-1}|x_n,x_0)$, is a denoising process that removes noise from an initialized noise state. At inference time, the model can generate an output, starting with noise and gradually removing it using $p_\theta$. 

In the discrete diffusion model, $q(x_n|x_{n-1}) := v^{T}(x_n)\textbf{Q}_{n}v(x_{n-1})$ and $p_\theta(x_{n-1}|x_n,y): =\sum^{k}_{\hat{x}_0=1}q(x_{n-1}|x_n,\hat{x_0})p_\theta(\hat{x_0}|x_n,x_0, y)$, where $v(x_n)$ is a one-hot vector with entry $1$ at $x_n$, and $\textbf{Q}_{n}$ is a transition matrix, modeling the probability to move from state $x_{n-1}$ to $x_n$, using uniform probability over the vocabulary and a pre-defined probability for additional special \textit{[MASK]} token.
$x_0$ is a discrete vector, tokenized by the VQGAN~\citep{esser2021taming} encoder. For modeling $p_\theta$ we have followed ~\citep{Gu2021VectorQD} and used a conditional Transformer ~\citep{vaswani2017attention}.\\
In the continuous diffusion model, $q(x_n | x_{n-1}) := \mathcal{N}(x_n; \sqrt{\alpha_t} x_{n-1}, (1-\alpha_n)x_0)$ and $p_{\theta}(x_{n-1}|x_n) := \mathcal{N}(\mu_{\theta}(x_n), \Sigma_{\theta}(x_n))$. Here, the noise function is Gaussian noise. For modeling $p_\theta$ we have followed ~\citep{ramesh2022hierarchical} and used a \textit{U-net} architecture.

We sample both our models using Classifier Free Guidance (CFG)~\citep{glide,cfg}. Since CFG was originally proposed for continuous models, we propose a method for using it with discrete models as well. The full details can be found in Sec.~\ref{cfg_supp} of the supplement.

\begin{figure}[t!]
\centering
\includegraphics[width=1.\linewidth]{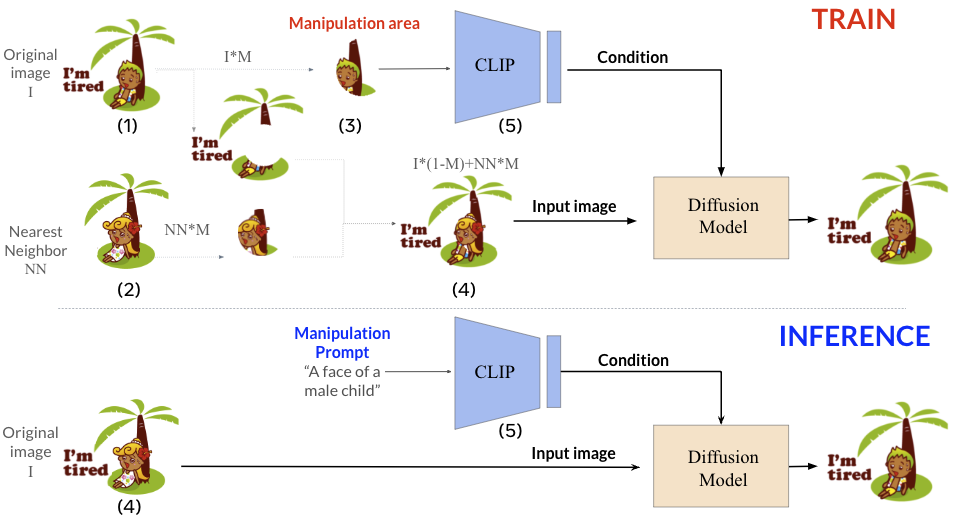}
\caption{An illustration of our manipulation approach. \textbf{During training:} Given a training image (1), the model extracts its first nearest neighbor (2). Next, a random local area in the training image is selected (3), and the manipulated image is constructed by replacing the area with the corresponding nearest neighbor (4). The model then receives as input the manipulated image and the clip embedding of the local area that needs to be restored (5). \textbf{During inference:} Given an input image and a text query "A face of a male child", the model receives as input the image (4) and the clip embedding of the modifying text (5).} 
\label{fig:manipulation_illustration}
\vspace{-0.4cm}
\end{figure}

\subsection{Text-only image manipulation}
The majority of previous works in the task of image manipulation either rely on user-provided masks~\citep{glide,avrahami2022blended, avrahami2022latent}, or are limited to global editing~\citep{crowson2022vqgan,kim2022diffusionclip}. Recently, several works~\citep{bar2022text2live,hertz2022prompt,gal2022image} have made progress with local manipulations without relying on user edited masks. Nevertheless, most of the techniques suffer from several shortcomings: (1) They enable local texture changes, yet cannot modify complex structures, (2) they struggle to preserve the identity of the object, for example, when manipulating humans, (3) they require optimization for each input.

We address these issues by extending \knndiffusion to perform local and semantic-aware image manipulations without any provided mask. Illustration of the approach is provided in Fig.~\ref{fig:manipulation_illustration} and in Fig.~\ref{fig:manipulations_construction} in the supplement. For this task, the model is trained to predict the original image from a manipulated version. Specifically, we create a manipulated version of the image, which differs from the original image only in some local area. Given a random local area $M$ in the image $\image$, the manipulated image $\image_{manip}$ is constructed by replacing the area with the corresponding nearest neighbor:
$\image_{manip}=\image\cdot(1-M)+\text{nn}_{img}(\image, 1)\cdot M$, where  $\text{nn}_{img}(\image, 1)$ is the the nearest neighbor obtained after aligning it with $\image$ using the ECC alignment algorithm~\citep{evangelidis2008parametric}. %
The model then receives as input the manipulated image, together with the \clip embedding of the original image \emph{only} in the local area: $\imageencoder(\image\cdot M)$. This \clip embedding represents the required modification that should be applied to the manipulated image in order to predict the original image.
During inference, instead of using the \clip embedding of the local area, the desired modification is represented using the \clip embedding of the user text query. %
We modified the model to be capable of receiving as a condition both the manipulated image and the \clip embedding of the local area.

\begin{figure}[t!]
\centering
\includegraphics[width=1.0\linewidth]{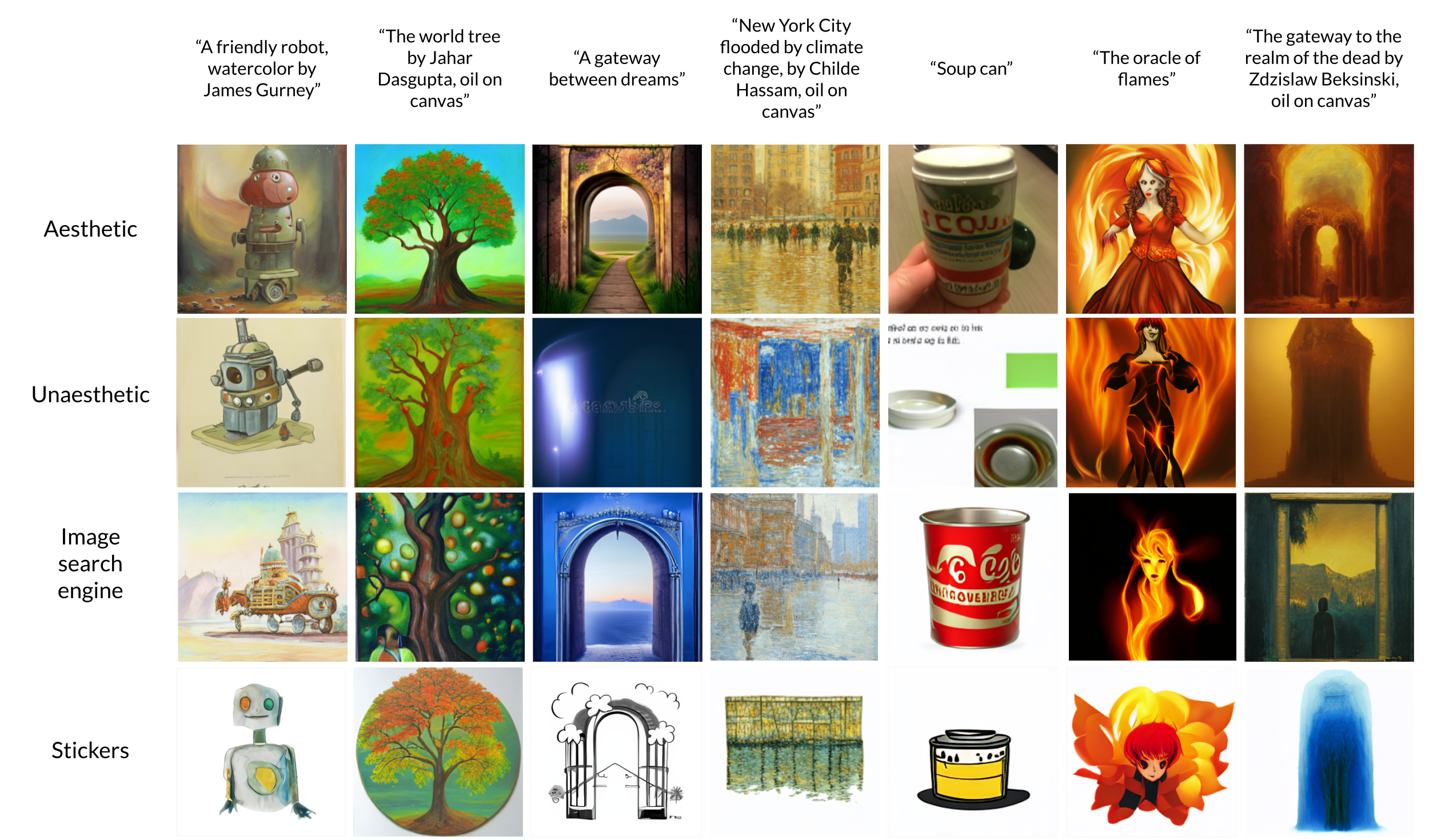} 
\caption{A qualitative comparison between different indexes used by the same model. In each row we have conditioned the model with \knn of a different index. (1) \textbf{Aesthetic.} Using images from the first quantile of an aesthetic classifier, (2) \textbf{Unaesthetic.} Using images from the last quantile of an aesthetic classifier, (3) \textbf{Image search engine.} Using images retrieved from Google Images, (4) \textbf{Stickers.} Using the stickers index.}
\label{fig:ood}
\vspace{-.4cm}
\end{figure}

\section{Experiments}
\label{section:experiments}
First, we conduct qualitative and quantitative comparisons on MS-COCO, LN-COCO and CUB datasets to compare with previous work. To further demonstrate the advantage of our method, we provide comparison on an image-only stickers dataset, where we apply our approach on two diffusion backbones. Next, we demonstrate image manipulation and out-of-distribution capabilities. Finally, to better assess the effect of each contribution, an ablation study is provided.  
\paragraph{Datasets and Metrics.} 
\label{subsection_datasets_and_metrics}
For photo-realistic experiments, our model was trained only on the images (omitting the text) of a modified version of the Public Multimodal Dataset~(PMD) used by FLAVA~\citep{singh2021flava}. More information about the dataset is available in Sec.~\ref{section_supp:datasets} of the supplement. %
To further demonstrate the capabilities of our method, we collected 400 million sticker images from the web, containing combinations of concepts such as objects, characters/avatars and text. 
The collected stickers do not have paired text, and are substantially different from photo-realistic data. Furthermore, since they have no paired text, they were not part of \clip's training data, which makes the text-to-image generation task more challenging.

Evaluation metrics are based on objective and subjective metrics: (i) FID~\citep{heusel2017gans} is an objective metric used to assess the quality of synthesized images, (ii) human evaluation - we ask human raters for their preference, comparing two methods based on image quality and text alignment. We used 600 image pairs; five raters rated each pair. The results are shown as a percentage of majority votes in favor of our method over the baselines. We report the full human evaluation protocol in the supplement. We chose to omit Inception-Score, since it is shown by \cite{barratt2018note} to be a misleading metric for models that were not trained on Imagenet.

\begin{figure*}[t!]
\centering
\includegraphics[width=1.0\linewidth]{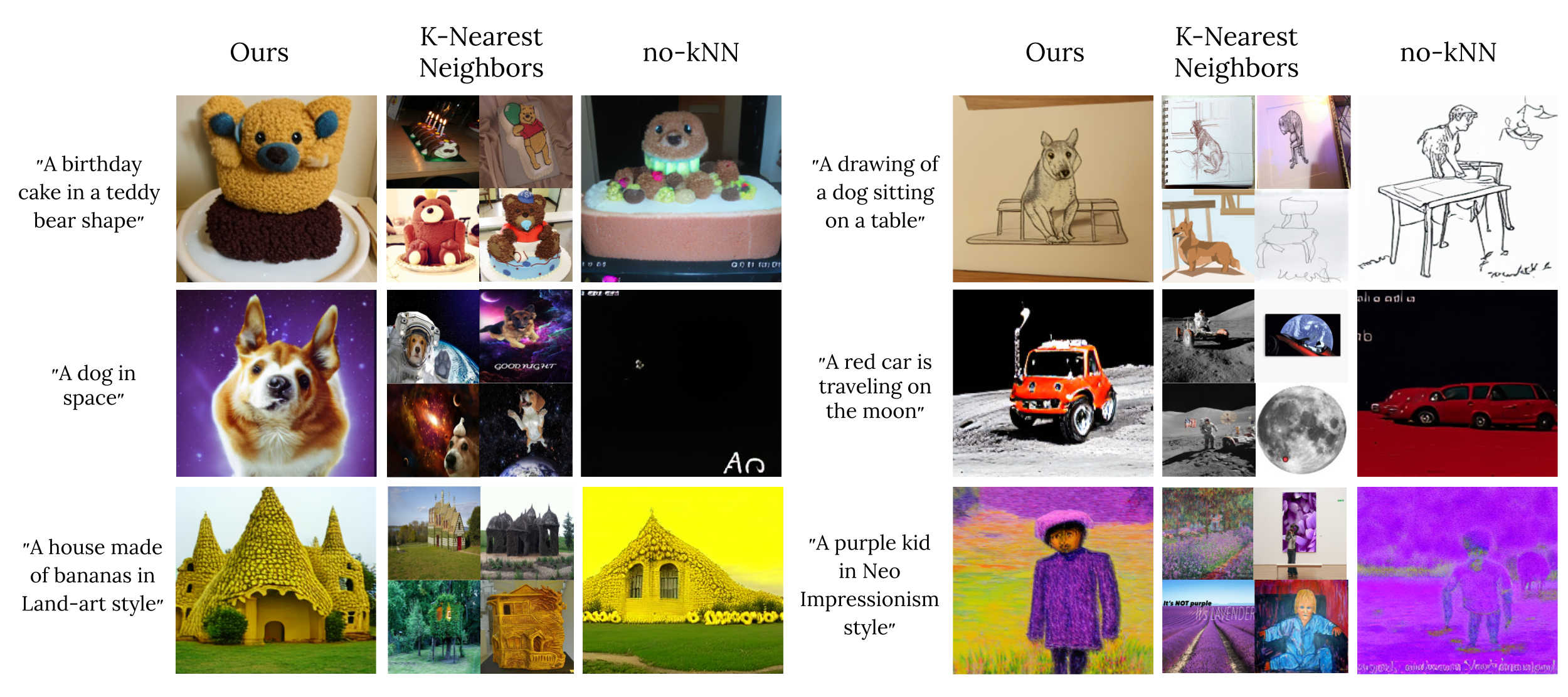} 
\caption{Comparison of our model, trained on PMD with (1) \knn extracted in inference, (2) the same model without using \knn in inference. While the \knn lack information regarding text semantics, our model considers both text semantics and the \knn, thus proving the advantage of using both the text and the \knn embeddings.}
\label{fig:cmd}
\vspace{-.4cm}
\end{figure*}

\subsection{Qualitative and Quantitative Results}
We begin by comparing our model, trained on the PMD dataset, with the previous works LAFITE and FuseDream, that trained on image-only datasets. %
To demonstrate the advantage of using a retrieval method in text-to-image generation, we trained a model variant, \noknn. This baseline was trained solely on image embeddings (omitting the \knn), while during inference, the images were generated using the text embedding. 
Tab.~\ref{tab:zero_shot_coco} displays zero-shot results on three different datasets: MS-COCO, CUB and LN-COCO. We follow the evaluation protocol of LAFITE, reporting our results on MS-COCO without training, nor using it's training partition in the \knn index. Similarly, we follow LAFITE for CUB and LN-COCO evaluation. As can be seen, our model achieves the lowest FID score in all scenarios. In addition, human evaluations rate our method as better aligned to text and with the highest images quality. In Fig.~\ref{fig:cmd_hard},~\ref{fig:cmd} and~\ref{fig:coco2} we present a qualitative comparison between the methods. 
One can observe that while the simple retrieval baseline outputs non-generated images with high-quality, the images generated by our method are more faithful to the input text.

To further demonstrate the effectiveness of our method, we present in Fig.~\ref{all_models_fid_coco} a comparison of our model with the latest text-to-image models trained on paired text-image datasets: DALL$\cdot$E, CogView, VQ-Diffusion, GLIDE, Latent Diffusion(LDM), Make-A-Scene, DALL$\cdot$E2, Parti and Imagen. As can be seen, our model achieves comparable results to recent models trained with full text-image pairs (e.g LDM, GLIDE), despite being trained on an image-only dataset, with significantly lower computational costs. The results demonstrate that leveraging an external retrieval database allows our method to compensate for different trade-offs, in particular, reducing the number of parameters in the model. Additional samples are provided in Fig.~\ref{fig:cmd_challanging} in the supplement. \\

\setlength{\tabcolsep}{4pt}
\begin{table}[t!]
\caption{Results on the stickers dataset. For both the continuous and the discrete models, we report the percentage of human raters prefer our method over the baselines with respect to image quality and text alignment. Discrete~\textit{no-\knn} refers to VQ-diffusion, trained without text, and Continuous ~\textit{no-\knn}, to DALL$\cdot$E2 decoder, trained without text.}
\vspace{-0.4cm}
\begin{center}
\scalebox{0.9}{
\begin{tabular}{l|c|cc|cc}
\toprule
  & &\multicolumn{2}{c}{Ours Discrete} & \multicolumn{2}{|c}{Ours Continuous}  \\
Model & FID$\downarrow$ & \begin{tabular}{@{}c@{}}Image quality\end{tabular}& \begin{tabular}{@{}c@{}}Text alignment\end{tabular} &  \begin{tabular}{@{}c@{}}Image quality\end{tabular}& \begin{tabular}{@{}c@{}}Text alignment\end{tabular}  \\
\noalign{\smallskip}
\hline
\noalign{\smallskip}
\begin{tabular}{@{}c@{}}DALL$\cdot$E2+ClipCap\end{tabular}  & $55.5$ & $71.6$ & $69.2$ & $67.0$ & $68.3$  \\
LAFITE & $58.7$ & $63.5$ & $59.9$ &  $76.0$ & $71.2$  \\
\noknn & $52.7$  & $72.1$ & $67.6$ &$66.8$  & $69.4$\\
Ours & $\textbf{40.8}$ & - & - & - & - \\
\hline
\end{tabular}
}
\end{center}
\label{tab:stickers}
\end{table}
\vspace{-0.7cm}
\setlength{\tabcolsep}{1.4pt}

\paragraph{Text-to-sticker generation.}
\label{text_to_sticker_generation}
As the sticker dataset does not have paired text, and is substantially different from photo-realistic data, it allows us to illustrate the advantage of our model on an image-only dataset. A selection of stickers generated by our model is presented in Fig.~\ref{fig:alpha_stickers_main} and Fig.~\ref{fig:stickers_challanging},~\ref{fig:alpha_stickers_supp}.

To demonstrate the importance of using \knn on image-only datasets, we evaluate our approach on two diffusion backbones. To this end, we trained a continuous diffusion model ~\citep{ramesh2022hierarchical} and a discrete diffusion model ~\citep{Gu2021VectorQD}, both conditioned on the \knn image embeddings.
For each backbone, we compare our method with the following baselines: (1) \noknn - this baseline was trained using both the continuous and the discrete methods conditioned only on image \clip embedding, without using \knn. In the discrete case, we trained a VQ-diffusion model, while in the continuous case, we trained a re-implementation of DALL$\cdot$E2's decoder (without prior). (2) \textit{DALL$\cdot$E2+ClipCap} baseline - here, we first captioned the entire sticker dataset using ClipCap~\citep{mokady2021clipcap}, then trained  DALL$\cdot$E2 decoder on the captioned dataset. (3) LAFITE - we trained LAFITE language-free model on our stickers dataset using the authors' published code.

We present the results in Tab.~\ref{tab:stickers}. The FID is calculated over a subset of $3,000$ stickers, generated from the ClipCap captioned dataset.
As can be seen, our model achieves the lowest FID score. In addition, it outperforms all baselines in human evaluation comparison, using both continuous and discrete backbones. In particular, compared with the same model trained without \knn, our model achieves significantly higher favorability in both text alignment and image quality.

\subsection{Applications}

\paragraph{Text-only image manipulation.}
\label{manipulations_section}
We demonstrate the manipulation capabilities of our model in Fig.~\ref{fig:alpha_stickers_main},~\ref{fig:human_manipulation} and~\ref{fig:expression_manipulations}.
Furthermore, we qualitatively compare our model with Text2LIVE~\citep{bar2022text2live} and Textual Inversion~\citep{gal2022image}, using the authors' published code. Text2LIVE proposed generating an edit layer that is composed over the original input, using a generator trained for each training image. Textual Inversion utilized the pre-trained Latent Diffusion model to invert the input image into a token embedding. The embedding is then used to compose novel textual queries for the generative model.
Fig.~\ref{fig:human_manipulation} shows representative results, and the rest are included in Fig.~\ref{fig:text2live_compare} and~\ref{fig:textual_compare} in the supplement. 
In contrast to our model, baseline methods lack text correspondence or they do not preserve the identity of the object. Since Text2LIVE is optimized to perform local changes, it has the difficulty changing the structure of the object (e.g. the "raising his hand" example in Fig.~\ref{fig:human_manipulation}). Textual Inversion baseline changes the identity of the object because it struggles reconstructing the textual representation of the source image. Our model, on the other hand, can perform challenging manipulations that are aligned with the text, while preserving the object identity.

\begin{wrapfigure}{r}{5.0cm}
\vspace{-10pt}
\centering
\includegraphics[width=4.5cm]{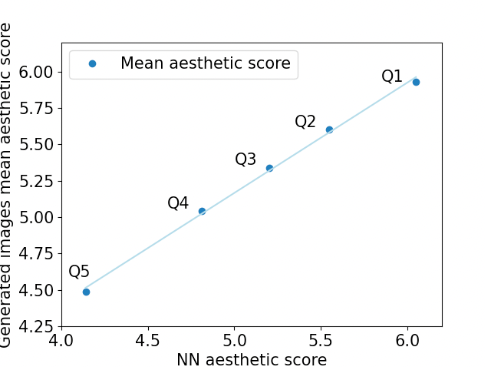}
\caption{Mean aesthetics score of the generated images as a function of the conditioned \knn mean aesthetics score. As can be seen, the generation can be controlled by filtering the \knn.}
\label{fig:aesthetics_classifier}
\vspace{5pt}
\end{wrapfigure} 
\paragraph{Out-of-distribution generation.}
\label{out_of_distribution_section}
Using the retrieval index as part of the generation process enables using different databases during inference, without fine-tuning. This allows the model to generate images from distributions that were not part of the training set, enabling out-of-distribution generation. This novel capability is demonstrated with the same model trained on PMD, using %
three different retrieval databases: \textit{(i) A stickers database} presented in Sec.~\ref{subsection_datasets_and_metrics}. (ii) \textit{Aesthetic database:} This database is constructed by filtering images according to a classifier score. Let $C$ be a classifier that for each image $i\in I$ outputs a score $s=C(i)$. This classifier enables filtering the \knn using $L \leq s < H$, where $L$ and $H$ are low and high thresholds, respectively. We demonstrate this capability using an open source pre-trained aesthetics classifier $A$ ~\citep{aesthetics_classifier}: For each text input $t\in T$, we apply $A$ on the \knn, and then divide the \knn into five equal quantiles based on $A$ score. As can be seen in Fig. ~\ref{fig:aesthetics_classifier}, using \knn with higher aesthetics score result in generated images with higher aesthetics mean score. \textit{(iii) Image search engine:} Generative models are stationary in the sense that they are unable to learn new concepts after being trained, hence fine-tuning is required to represent new styles and concepts. Here, we use an online image search engine, which allows the model to adapt to new data without additional fine-tuning. 
A qualitative comparison of all three methods is shown in Fig.\ref{fig:ood}.

\begin{wrapfigure}{r}{0.4\textwidth}
\centering
\includegraphics[width=0.35\textwidth]{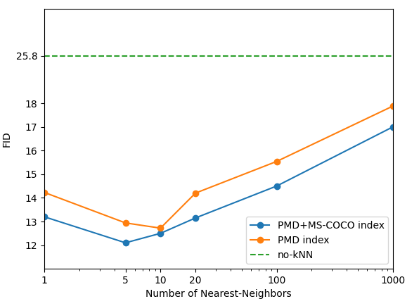}
\vspace{-10pt}
\caption{MS-COCO test FID score on various K's in two settings: (1) Zero-Shot (2) Index includes the MS-COCO training subset. \textbf{No index inference} was trained with \knn, but did not employ \knn in inference. 
}\label{fig:ablation_num_knn}
\end{wrapfigure} 

\subsection{Ablation Study}
We conclude our experiments with an ablation study, to quantify the contribution of our different components. The experiments are performed on the PMD dataset, and FID scores are reported on the MS-COCO dataset. Ablation analysis on index size and different \knn conditioning approaches are provided in Sec.~\ref{section_supp:ablation} in the supplement. 

\textbf{Number of nearest neighbors.} The results in Fig.~\ref{fig:ablation_num_knn} demonstrate the importance of applying the retrieval mechanism during training and inference. Here, we evaluate our \knndiffusion model, trained on PMD dataset, with different numbers of \knn during inference - $1,5, 10,20, 100,1000$. Furthermore, we examined the baseline \textit{no-\knn}, in which during inference, the model is conditioned only on the text input embedding $f_{txt}(t)$, without using \knn. As can be seen, best performance is achieved using $10$ neighbors.

\section{Conclusion}
\label{section:discussion}
\vspace{-.1cm}
``We shall always find, that every idea which we examine is copied from a similar impression", \cite{Hume}. 
In this paper, we propose using a large-scale retrieval method in order to train a novel text-to-image model, without any text data. Our extensive experiments demonstrate that using an external knowledge-base alleviates much of the model's burden of learning novel concepts, enabling the use of a relatively small model. In addition, it provides the model the capability of learning to adapt to new samples, which it only observes during test time. Lastly, we present a new technique utilizing the retrieval method for text-driven semantic manipulations without user-provided masks. As evaluated by human studies and automatic metrics, our method is significantly preferable to the baselines in terms of image quality and text alignment.

\bibliography{iclr2023_conference}
\bibliographystyle{iclr2023_conference}

\clearpage
\section{Appendix}

\subsection{Human Evaluation Protocol}
For all of our human evaluation experiments, we used Amazon Mechanical Turk. For each experiment, we used 600 samples, each scored by five different people. The preferred sample was determined according to majority opinion. For each baseline comparison, we asked two questions (in different experiments): "Which image is of a higher quality?" and "Which image best matches the text?".

\subsection{Datasets}
\label{section_supp:datasets}
The modified PMD dataset is composed of the following set of publicly available text-image datasets: SBU Captions~\citep{sbu_captions}, Localized Narratives~\citep{pont2020connecting}, Conceptual Captions~\citep{sharma2018conceptual}, Visual Genome~\citep{visualgenome}, Wikipedia Image Text~\citep{srinivasan2021wit}, Conceptual Captions 12M~\citep{cc12m}, Red Caps~\citep{redcaps}, and a filtered version of YFCC100M~\citep{yfcc100m}. In total, the dataset contains 69 million text-image pairs.

\subsection{Ablation study}
\label{section_supp:ablation}
\textbf{Index size} As one can expect, increasing the index size at inference time improves performance. To demonstrate this hypothesis, we evaluated our model with an index containing 10\%, 30\%, 50\% and 70\% images of PMD dataset, and obtained FID scores of $13.92$, $13.85$, $13.72$, and $13.65$ respectively. 

\textbf{\knn conditioning} We examined several different approaches to \knn input conditioning: (i) forwarding the \knn embeddings and the single image embedding through a self-attention layer before feeding the contextualized $K+1$ embeddings to the model, (ii) feeding the model with one embedding, computed using cross-attention between the image embedding and the \knn embeddings, and, (iii) feeding the model with the image embedding concatenated with a learned linear projection of the \knn embeddings. These variants received FID scores of 18.3, 22.4, 34.1 respectively.

\subsection{Retrieval Model}
\label{subsection_supp:retieval}
The retrieval model is implemented using FAISS \citep{faiss}. FAISS is an efficient database, capable of storing billions of elements and finding their nearest neighbors in milliseconds. In the pre-process phase we store the image index and its corresponding CLIP image embedding. Then, during training we search for the nearest neighbors using the image index.  

The photo-realistic dataset contains $69M$ samples, and the sticker dataset contains $400M$ samples. For an efficient search during training and inference, we use an inverted file index. As in~\citet{babenko2014inverted}, we define Voronoi cells in the $d$-dimensional space (where $d=512$ is the CLIP embedding dimensional space), s.t each database vector falls in one of the cells. When searching for an image $I$, the nearest neighbors are retrieved only from the cluster that $I$ falls in, and from the 19 other neighboring clusters.
In order to fit the index of our large-scale datasets on a 128GB RAM server, we compress the code size from $4d=2048$ to $256$ using optimized product quantization~\citep{Ge_2013_CVPR, jegou2010product}. 

\subsection{Discrete KNN model}
\label{subsection_supp:discrete}
We provide additional implementation details for the discrete diffusion model.

\paragraph{Vector Quantization}
For token quantization, we use VQ-VAE and adapt the publicly available VQGAN\citep{esser2021taming} model, trained on the OpenImages\citep{openimages} dataset. The encoder downsamples images to $32 \times 32$ tokens and uses a codebook vocabulary with $2887$ elements.

\paragraph{Image Tokenization}
In our discrete generative model we model images as a sequence of discrete tokens. To this end, we utilize a vector-quantized variational auto-encoder~(VQ-VAE)~\citep{van2017neural} as image tokenizer. VQ-VAE consists of three components: (i) an encoder, (ii) a learned codebook, and, (iii) a decoder. Given an image, the encoder extracts a latent representation. The codebook then maps each latent vector representation to its nearest vector in the codebook. Finally, the decoder reconstructs the image from the codebook representation. VQ-VAE is trained with the objectives of reconstruction and codebook learning. VQ-GAN~\citep{esser2021taming} adds an adversarial loss term that tries to determine whether the generated image is fake or real. This added term was shown to improve reconstruction quality.

\paragraph{Transformer}
We follow~\cite{Gu2021VectorQD} and train a decoder-only Transformer. 
The decoder module contains 24 transformer blocks, each containing full attention, cross-attention for the concatenated conditioner, and a feed-forward network. The timestamp $n$ is injected using Adaptive Layer Normalization \citep{ba2016layer}. The decoder contains 400 million parameters, which is merely a tenth of baseline models such as CogView, DALL$\cdot$E, and GLIDE. We use the same architecture for both sticker and photo-realistic image generation.

\paragraph{Classifier-free guidance}
\label{cfg_supp}
We sample our diffusion models using classifier-free guidance~(CFG)~\citep{cfg,glide,ramesh2022hierarchical}. 
CFG is performed by extrapolating an unconditional sample in the direction of a conditional sample. To support unconditional sampling, previous work had to fine-tune~\citep{glide} their models with 20\% of the conditional features nullified. This enabled them to sample unconditional images from the model using the null condition, $y'=\overrightarrow{0}$, the null vector. We found that we can generate unconditional samples from our model using null conditioning without fine-tuning it. We hypothesize that by conditioning the model on a null vector, the cross-attention component is also nullified, resulting in no contribution to the diffusion process.
During inference, in each step of the diffusion process we generate two images: conditional image logits, $p_\theta(x_{n-1} | x_n, y)$, conditioned on the desired multi-modal embedding $y$, and the unconditional image logits, $p_\theta(x_{n-1} | x_n, y')$, conditioned on the null embedding. Then, the final image for a diffusion step $n$ is sampled from
\begin{align*}
    p_\theta(x_{n-1} | x_n, y) = & p_\theta(x_{n-1} | {x_n, y'}) + \\ & \lambda(p_\theta(x_{n-1} | x_n, y) - p_\theta( x_{n-1} | x_n, y'))
\end{align*} where $\lambda$ is a scale coefficient. In all of our experiments, we set $\lambda=8$, which was found to yield the highest FID scores on the validation set. Note that the above extrapolation occurs directly on the logits output by $p_\theta$, in contrast to GLIDE~\citep{glide}, which extrapolates the pixel values. 

\paragraph{Training Objective}
\label{supp_discrete_training_objective}
For completeness we are adding the training objective of the discrete model. The network is trained to minimize the variational lower bound (VLB): 
\begin{equation}
\begin{split}
\mathcal{L}_{\mathrm{vlb}} &= \mathcal{L}_0 + \mathcal{L}_1 + \cdots + \mathcal{L}_{N-1} + \mathcal{L}_{N}, \\ 
\mathcal{L}_0 &= -\log p_\theta(\bm{x}_0|\bm{x}_1,\bm{y}), \\ 
\mathcal{L}_{n-1} &= \kld{q(\bm{x}_{n-1}|\bm{x}_n,\bm{x}_0)}{p_\theta(\bm{x}_{n-1}|\bm{x}_n,\bm{y})}, \\
\mathcal{L}_N &= \kld{q(\bm{x}_N|\bm{x}_0)}{p(\bm{x}_N)}
\end{split}
\label{eqn:L_vlb}
\end{equation}
Where $p(\bm{x}_N)$ is the prior distribution of timestep $N$. In all of our experiments $N=1000$. The full details can be found in \citet{Gu2021VectorQD}.

\paragraph{Algorithms}
We provide the training algorithm ~\ref{algo:knndiff} and the inference algorithm ~\ref{algo:knndifftest} for the discrete diffusion model.\\

\begin{algorithm}
    \caption{Training of KNN-Diffusion}\label{algo:knndiff}
    \begin{algorithmic}[1]
        \STATE {\bfseries Input: A training image dataset $\mathcal{I}$, pre-trained image encoder $f_{img}$, image tokenizer $E$, learning rate $\eta$, maximum number of diffusion steps N, and loss weighting coefficient $\xi$}
        \STATE Store all dataset embeddings $\{\imageencoder(I_{i}) | i \in \database\} $ in a data-structure $\index$ 
    \REPEAT
        \STATE Sample an image $I$ 
        \STATE {\color{blue} // Get K nearest neighbors: } \\
        {\bfseries $\text{knn}_{img}(\image, k):=\argmin^k_{h \in \index}\dist(\imageencoder(\image), h)$}
        \STATE $y = (\imageencoder(I), \text{knn}_{img}(\image, k))$ {\color{blue} // Concatenate} 
		\STATE $x_0 = E(I)$ {\color{blue} // Get tokens from VQGAN-Encoder}
		\STATE Sample time step: $n \sim \text{Uniform}(\{1, \cdots, N\})$
		\STATE Sample $x_n$ from the forward process $q(x_n|x_0)$
		\IF{n=1}
		    \STATE $\mathcal{L} = \mathcal{L}_0$
		\ELSE
		    \STATE $\mathcal{L} = \mathcal{L}_{n-1} + \xi \mathcal{L}_{x_0}$
		\ENDIF
		\STATE $\theta \gets \theta - \eta \nabla_\theta \mathcal{L}$
    \UNTIL{converged}\\
    \end{algorithmic}
\end{algorithm}
\vspace{15px}

\begin{algorithm}
    \caption{Inference with KNN-Diffusion}\label{algo:knndifftest}
    \begin{algorithmic}[1]
        \STATE {\bfseries Input: Query text $t$, a dataset $\index$ of pre-processed image embeddings, pre-trained text encoder $\textencoder$, image de-tokenizer decoder \textbf{D}} 
        \STATE {\color{blue} // Get K nearest neighbors: }
        \STATE {\bfseries $\text{knn}_{txt}(\text, k):=\argmin^k_{h \in \index}\dist(\textencoder(t), h)$) }
        \STATE {$y = (\textencoder(t), \text{knn}_{txt}(\text, k))$ \color{blue} // Concatenate}
        
		\STATE {$x_n \gets [MASK]^{h \times w}$ \color{blue} // Initialize with mask tokens}
		\STATE $n \gets N$
		\WHILE {{\em $n > 0$}}{
    		\STATE $x_{n-1} \gets  \text{sample from}~p_\theta(x_{n-1}|x_n, y)$ 
    		\STATE {$n \gets n - 1$ }
		}
		\ENDWHILE
		\STATE {$I'$ = \textbf{D}$(x_{0})$}
    \end{algorithmic}
\end{algorithm}

\begin{figure}[t]
   \centering
\begin{tabular}{@{\hspace{-40\tabcolsep}}c@{\hspace{0.5\tabcolsep}}c@{\hspace{0.5\tabcolsep}}c@{\hspace{1.\tabcolsep}}c@{\hspace{1.\tabcolsep}}c@{\hspace{1.\tabcolsep}}c@{\hspace{1.\tabcolsep}}c@{\hspace{1.\tabcolsep}}c}

\resizebox{!}{11px}{
\begin{tabular}[x]{@{}c@{}}A black cat\\ with a beanie\\ on it's head\end{tabular}}&
\raisebox{-.5\height}{
\includegraphics[width=\imsize]{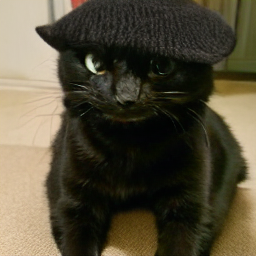}}&
\raisebox{-.5\height}{
\includegraphics[width=\imsize]{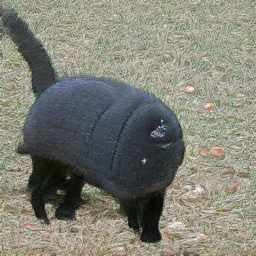}}&
\raisebox{-.5\height}{
\includegraphics[width=\imsize]{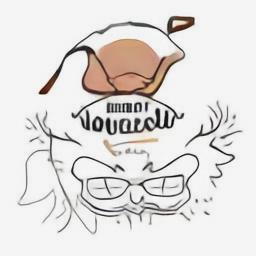}}&
\raisebox{-.5\height}{
\includegraphics[width=\imsize]{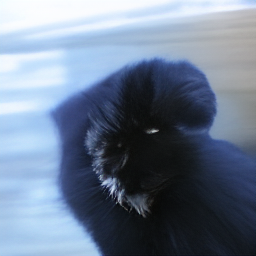}}&
\raisebox{-.5\height}{
\includegraphics[width=\imsize]{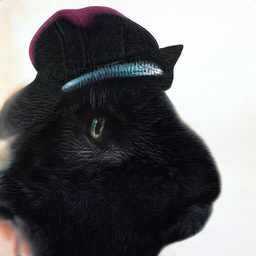}}&
\raisebox{-.5\height}{
\includegraphics[width=\imsize]{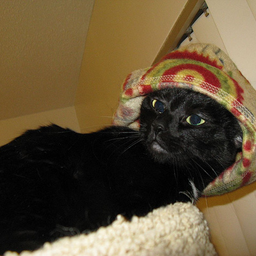}}&
\raisebox{-.5\height}{
\includegraphics[width=\imsize]{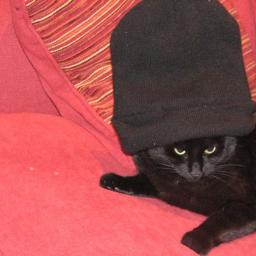}}\\

\resizebox{!}{11px}{
\begin{tabular}[x]{@{}c@{}}A bike is on\\ the street with\\ a lot of snow\end{tabular}}&
\raisebox{-.5\height}{
\includegraphics[width=\imsize]{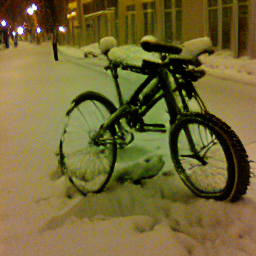}}&
\raisebox{-.5\height}{
\includegraphics[width=\imsize]{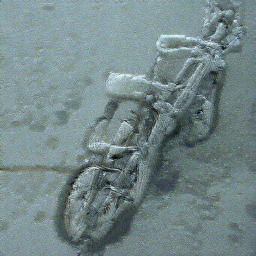}}&
\raisebox{-.5\height}{
\includegraphics[width=\imsize]{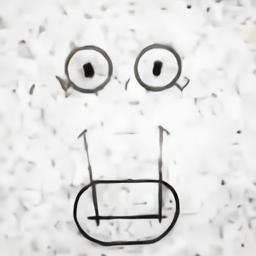}}&
\raisebox{-.5\height}{
\includegraphics[width=\imsize]{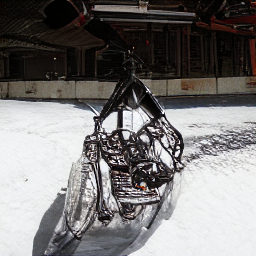}}&
\raisebox{-.5\height}{
\includegraphics[width=\imsize]{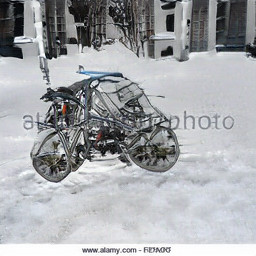}}&
\raisebox{-.5\height}{
\includegraphics[width=\imsize]{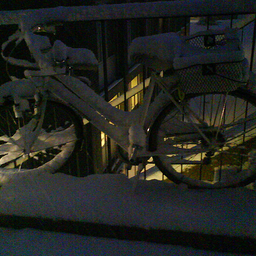}}&
\raisebox{-.5\height}{
\includegraphics[width=\imsize]{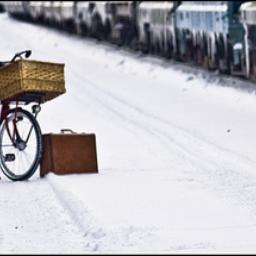}}\\

\resizebox{!}{11px}{
\begin{tabular}[x]{@{}c@{}}This is a \\living room in\\ a log cabin\end{tabular}}&
\raisebox{-.5\height}{
\includegraphics[width=\imsize]{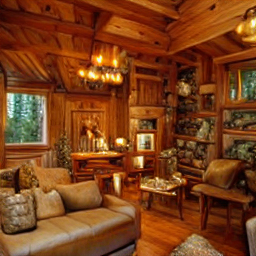}}&
\raisebox{-.5\height}{
\includegraphics[width=\imsize]{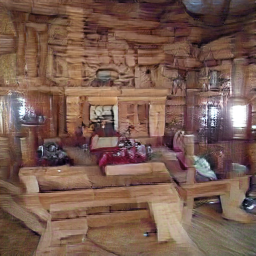}}&
\raisebox{-.5\height}{
\includegraphics[width=\imsize]{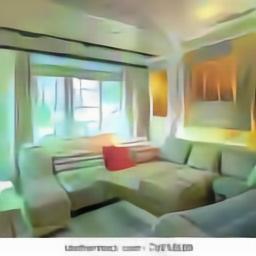}}&
\raisebox{-.5\height}{
\includegraphics[width=\imsize]{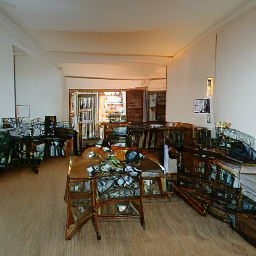}}&
\raisebox{-.5\height}{
\includegraphics[width=\imsize]{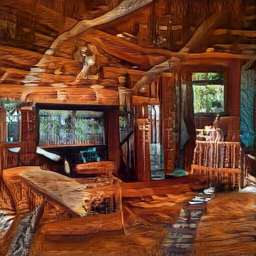}}&
\raisebox{-.5\height}{
\includegraphics[width=\imsize]{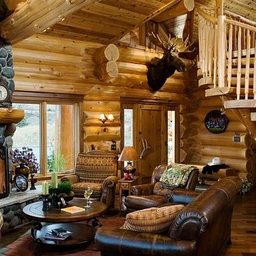}}&
\raisebox{-.5\height}{
\includegraphics[width=\imsize]{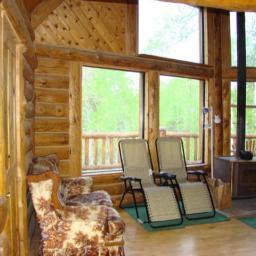}}\\

\resizebox{!}{12px}{
\begin{tabular}[x]{@{}c@{}}A neat and \\clean bathroom \\ in blue and\\ white\end{tabular}}&
\raisebox{-.5\height}{
\includegraphics[width=\imsize]{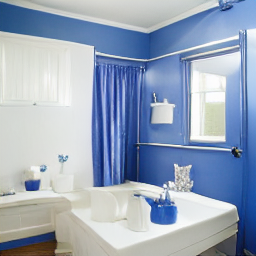}}&
\raisebox{-.5\height}{
\includegraphics[width=\imsize]{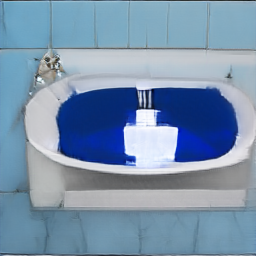}}&
\raisebox{-.5\height}{
\includegraphics[width=\imsize]{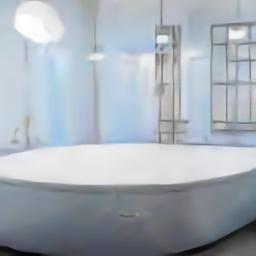}}&
\raisebox{-.5\height}{
\includegraphics[width=\imsize]{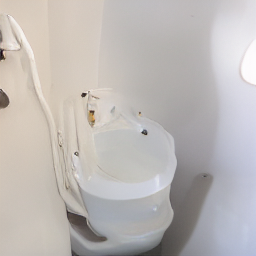}}&
\raisebox{-.5\height}{
\includegraphics[width=\imsize]{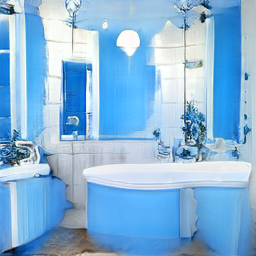}}&
\raisebox{-.5\height}{
\includegraphics[width=\imsize]{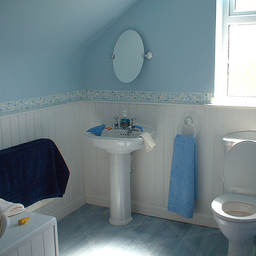}}&
\raisebox{-.5\height}{
\includegraphics[width=\imsize]{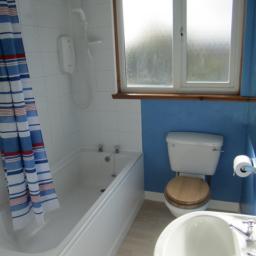}}\\

\resizebox{!}{12px}{
\begin{tabular}[x]{@{}c@{}}A knife and fork\\ on a plate\\ of pizza topped\\ with arugula\end{tabular}}&
\raisebox{-.5\height}{
\includegraphics[width=\imsize]{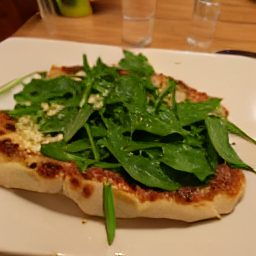}}&
\raisebox{-.5\height}{
\includegraphics[width=\imsize]{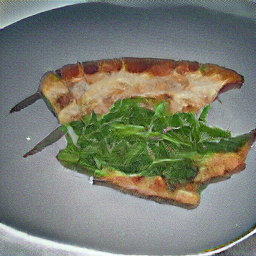}}&
\raisebox{-.5\height}{
\includegraphics[width=\imsize]{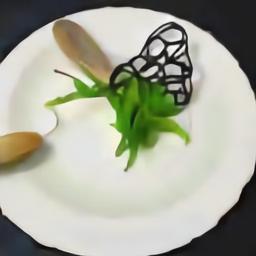}}&
\raisebox{-.5\height}{
\includegraphics[width=\imsize]{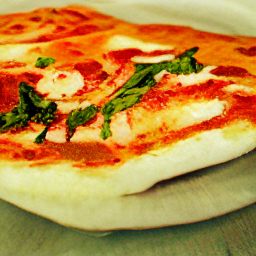}}&
\raisebox{-.5\height}{
\includegraphics[width=\imsize]{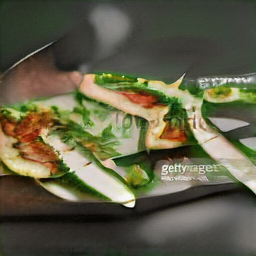}}&
\raisebox{-.5\height}{
\includegraphics[width=\imsize]{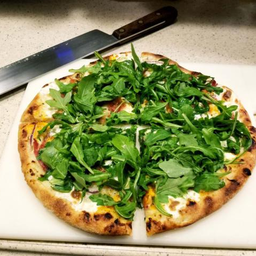}}&
\raisebox{-.5\height}{
\includegraphics[width=\imsize]{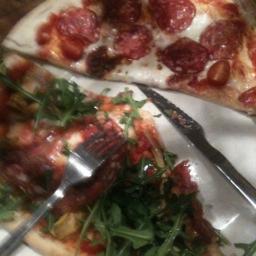}}\\

\resizebox{!}{11px}{
\begin{tabular}[x]{@{}c@{}}A lovely\\ appointed kitchen\\ area with oval\\ shaped island\end{tabular}}&
\raisebox{-.5\height}{
\includegraphics[width=\imsize]{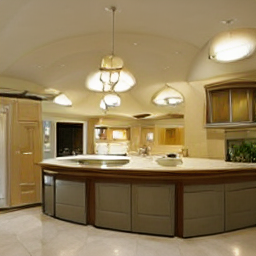}}&
\raisebox{-.5\height}{
\includegraphics[width=\imsize]{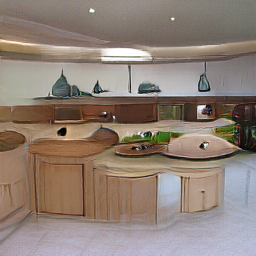}}&
\raisebox{-.5\height}{
\includegraphics[width=\imsize]{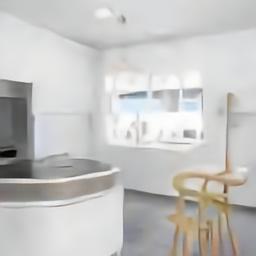}}&
\raisebox{-.5\height}{
\includegraphics[width=\imsize]{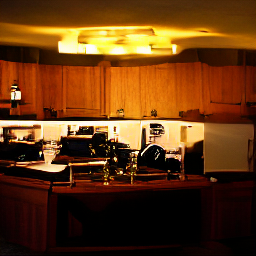}}&
\raisebox{-.5\height}{
\includegraphics[width=\imsize]{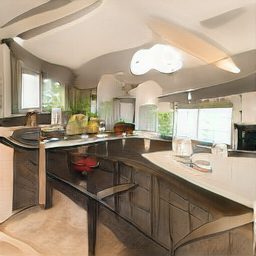}}&
\raisebox{-.5\height}{
\includegraphics[width=\imsize]{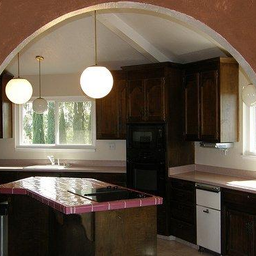}}&
\raisebox{-.5\height}{
\includegraphics[width=\imsize]{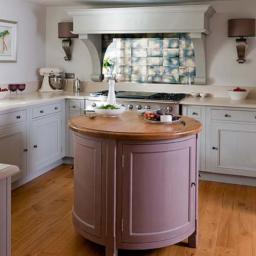}}\\

\resizebox{!}{11px}{
\begin{tabular}[x]{@{}c@{}}Rusty fire \\ hydrant is close \\ to the edge of \\ the curb and \\ painted green.\end{tabular}}&
\raisebox{-.5\height}{
\includegraphics[width=\imsize]{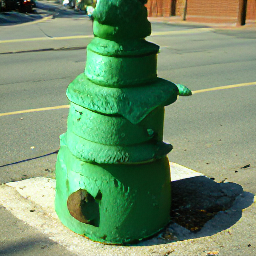}}&
\raisebox{-.5\height}{
\includegraphics[width=\imsize]{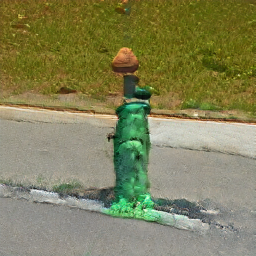}}&
\raisebox{-.5\height}{
\includegraphics[width=\imsize]{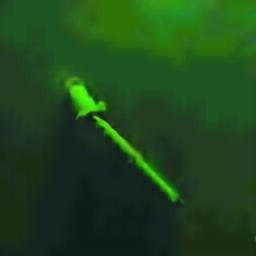}}&
\raisebox{-.5\height}{
\includegraphics[width=\imsize]{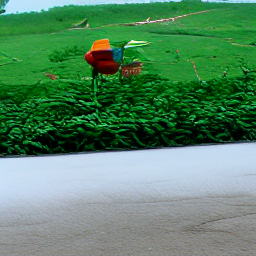}}&
\raisebox{-.5\height}{
\includegraphics[width=\imsize]{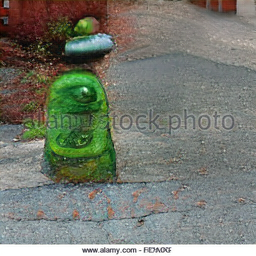}}&
\raisebox{-.5\height}{
\includegraphics[width=\imsize]{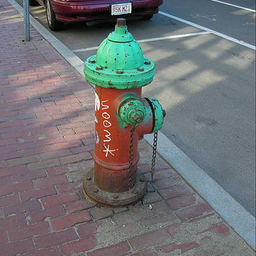}}&
\raisebox{-.5\height}{
\includegraphics[width=\imsize]{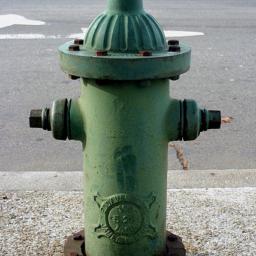}}\\

\resizebox{!}{12px}{
\begin{tabular}[x]{@{}c@{}}One cake \\ doughnut and \\ one doughnut \\topped with \\ icing\end{tabular}}&
\raisebox{-.5\height}{
\includegraphics[width=\imsize]{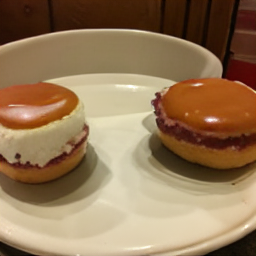}}&
\raisebox{-.5\height}{
\includegraphics[width=\imsize]{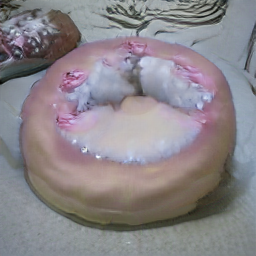}}&
\raisebox{-.5\height}{
\includegraphics[width=\imsize]{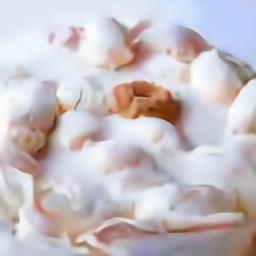}}&
\raisebox{-.5\height}{
\includegraphics[width=\imsize]{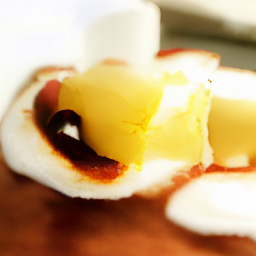}}&
\raisebox{-.5\height}{
\includegraphics[width=\imsize]{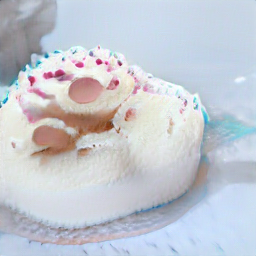}}&
\raisebox{-.5\height}{
\includegraphics[width=\imsize]{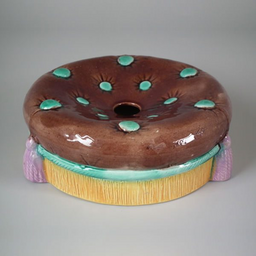}}&
\raisebox{-.5\height}{
\includegraphics[width=\imsize]{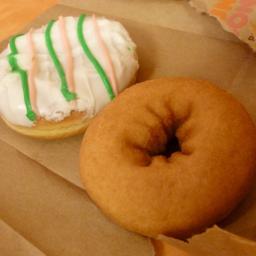}}\\

\resizebox{!}{11px}{
\begin{tabular}[x]{@{}c@{}}Tiered wedding\\ cake decorated \\with ribbon\\and daisies\end{tabular}}&
\raisebox{-.5\height}{
\includegraphics[width=\imsize]{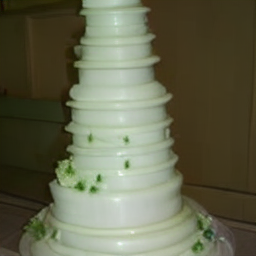}}&
\raisebox{-.5\height}{
\includegraphics[width=\imsize]{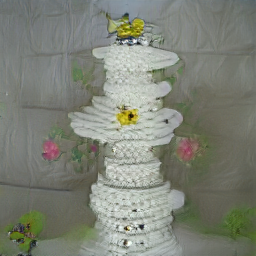}}&
\raisebox{-.5\height}{
\includegraphics[width=\imsize]{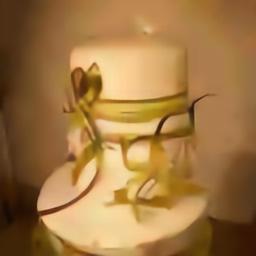}}&
\raisebox{-.5\height}{
\includegraphics[width=\imsize]{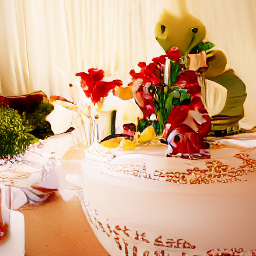}}&
\raisebox{-.5\height}{
\includegraphics[width=\imsize]{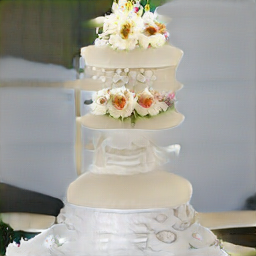}}&
\raisebox{-.5\height}{
\includegraphics[width=\imsize]{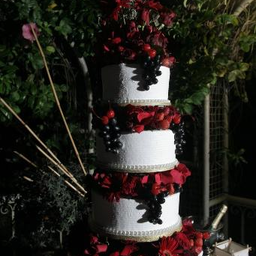}}&
\raisebox{-.5\height}{
\includegraphics[width=\imsize]{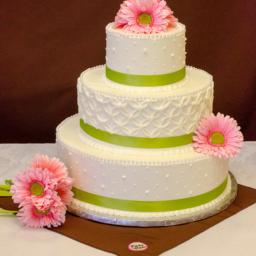}}\\

& Ours & \begin{tabular}[x]{@{}c@{}}FuseDream\end{tabular}  & \begin{tabular}[x]{@{}c@{}}CogView \end{tabular} & \begin{tabular}[x]{@{}c@{}}VQ \\  Diffusion\end{tabular} &\begin{tabular}[x]{@{}c@{}}Lafite \end{tabular} & NN & GT\\
\end{tabular}

    \caption{Samples from COCO validation set.}
    \label{fig:coco2} %
\end{figure}

\paragraph{COCO Validation Set Comparison} Fig.~\ref{fig:coco2} presents a qualitative comparison with FuseDream~\citep{fusedream}, CogView~\citep{cogview} and VQ-Diffusion~\citep{Gu2021VectorQD} on the COCO validation set. Note that both CogView and VQ-Diffusion have been trained on an Image-Text \emph{paired} dataset, whereas our model was not trained on the COCO dataset, nor used it in the retrieval model.

\begin{figure*}[ht!]
\centering
\includegraphics[width=1\linewidth]{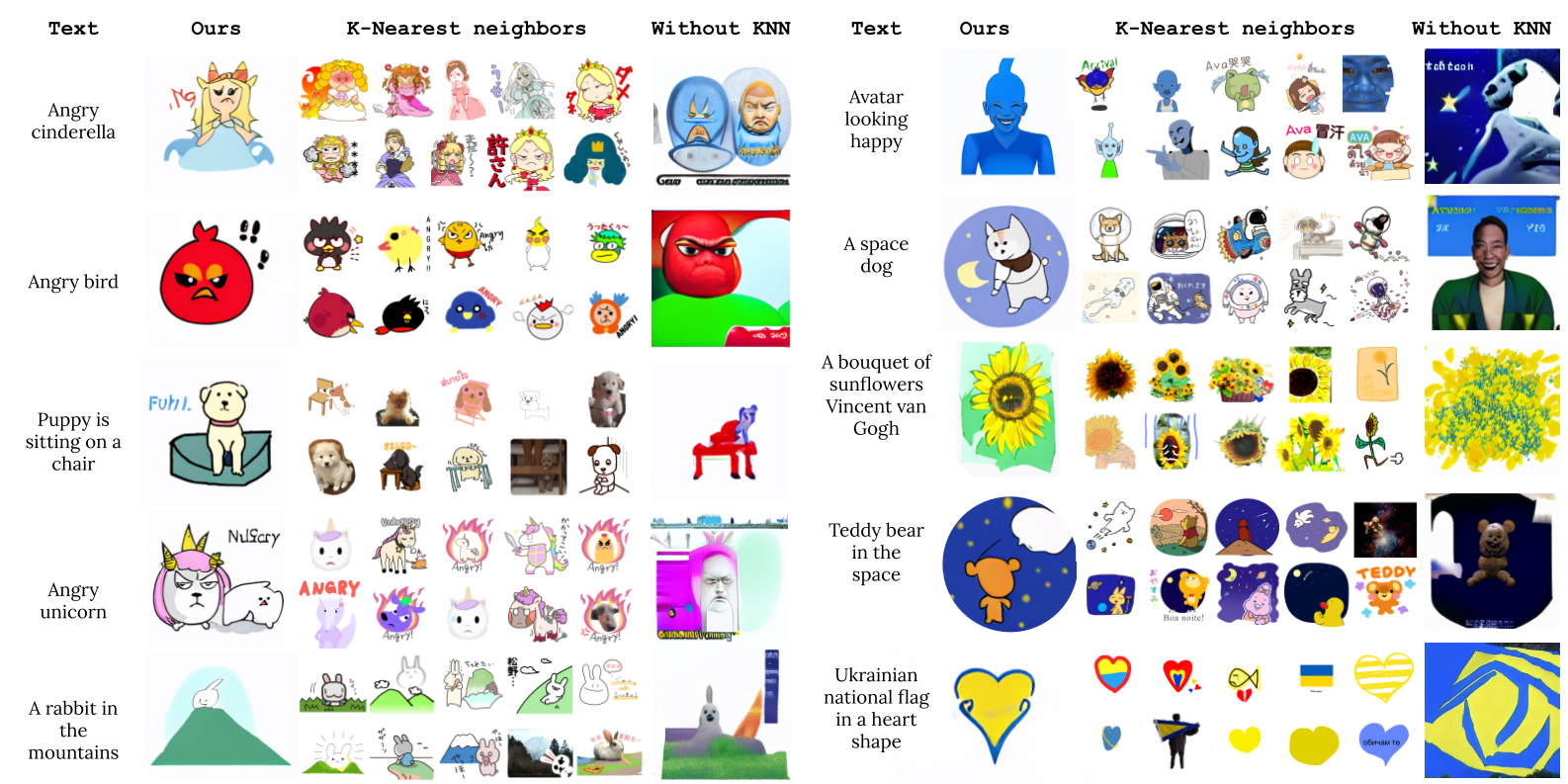} 
\caption{Qualitative comparison of stickers generated using the discrete \knndiffusion model, 10 Nearest Neighbors to the text in the CLIP embedding and a discrete model that does not use \knn.} 
\label{fig:stickers}
\end{figure*}

\paragraph{Sticker generation} 
In Fig.~\ref{fig:stickers} we present a visual comparison of our discrete model, trained on the stickers dataset with (1) the \knn extracted during inference, (2) the same model without using \knn in inference. As can be seen, the stickers generated by our model are better aligned to the corresponding text compared to the baselines. While the baselines fail with challenging prompts, our model produces high-quality images that align with the text, and composes multiple concepts correctly. 

\subsection{Continuous KNN model}
\label{subsection_supp:continuous}
 
\paragraph{Decoder.} We followed ~\citep{glide,ho2020denoising,ramesh2022hierarchical} and re-implemented a diffusion \textit{U-net} model. Specifically, we modify the architecture described in~\citep{ramesh2022hierarchical} by allowing multiple \clip embeddings as the condition to the model. Since we do not have a paired text-image dataset, we removed the text transformer, and thus the text embedding. Similarly to our discrete model, we trained two models (1) a \noknn conditioned only on \clip image embedding during training, (2) a \knn conditioned on \clip image embedding and its \knn. Finally, we enable classifier-free guidance by randomly setting the \clip embeddings to zero 10\% of the time. As demonstrated in Tab.~\ref{tab:stickers}, we find that humans prefer our model over \noknn $66.8\%$ of the time for image quality and $69.4\%$ of the time for text alignment.

\paragraph{Super-Resolution.} As the decoder generates images with $64\times64$ resolution, we up-sampled the images to $256\times256$ using the open-source super resolution of~\citep{glide}. To further up-sample the images to $512\times512$ and $1024\times1024$ we used the open-source super resolution provided by~\citep{wang2021real}.

\paragraph{Training Objectives}
For completeness we are addding the training objective of our continuous model. Following \citet{ho2020denoising, glide} we are using mean-squared
error loss to predict the noise:
$$L := E_{n \sim [1,N],x_0 \sim q(x_0), \epsilon \sim \mathcal{N}(0, \mathbf{I})}[||\epsilon - \epsilon_{\theta}(x_n, n, y)||^2]$$ where $\epsilon_{\theta}$ is a $U-net$ model and $y$ is the condition for the model. In all of our experiments $N=1000$.

\subsection{Text-only image manipulation}
\begin{figure}[ht!]
\centering
\includegraphics[width=1.\linewidth]{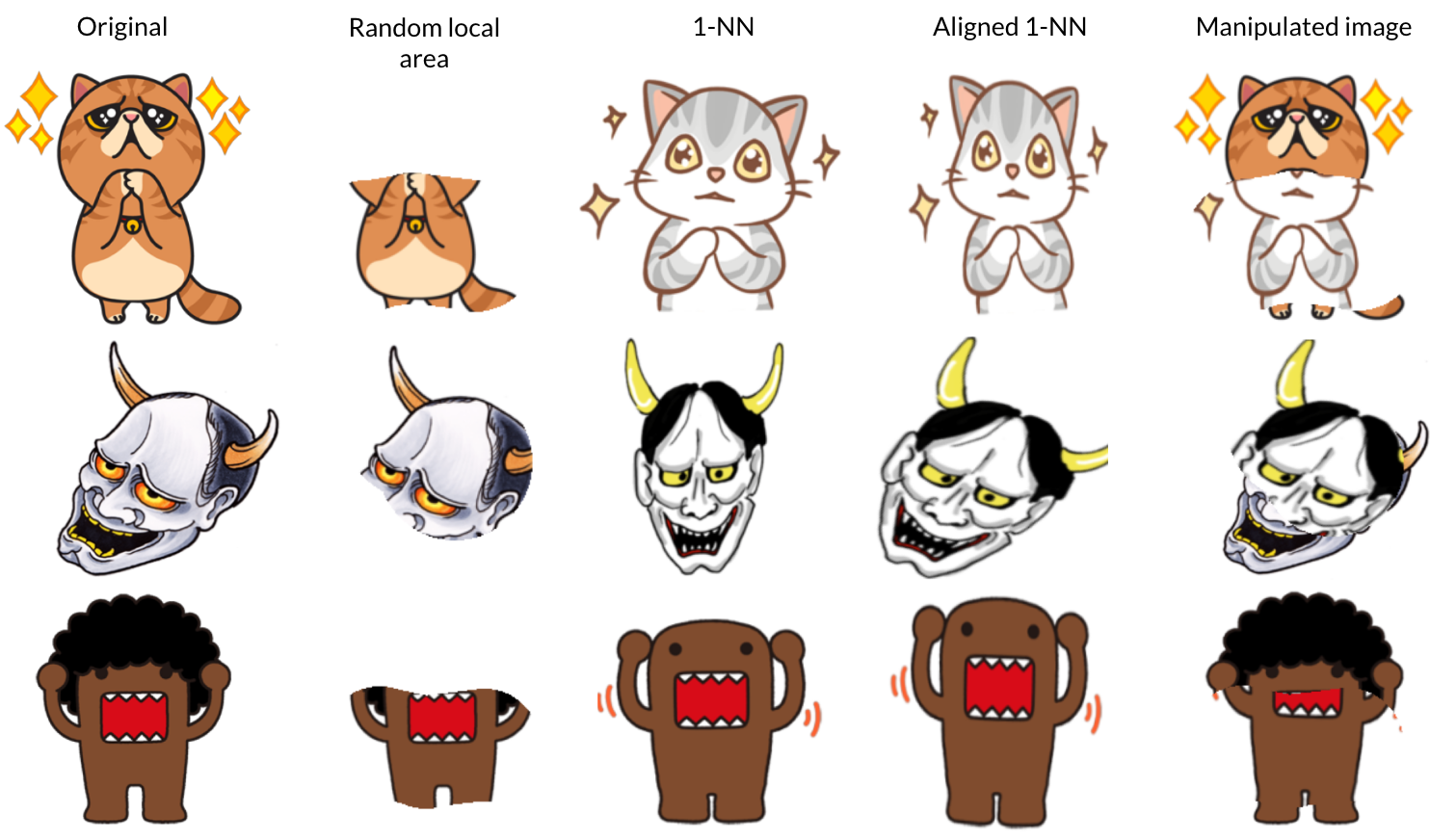} 
\caption{Illustration of the manipulated image construction process during training.
Given an original image, we select a random local area, and extract the first nearest neighbor (1-NN). Using ECC alignment, we align the nearest neighbor with the original image and replace the random local area with it's corresponding nearest neighbor local area. The model then receives as input the manipulated image, together with the \clip embedding of the local area, and tries to predict the original image.} 
\label{fig:manipulations_construction}
\end{figure}

\begin{figure}[ht!]
\centering
\includegraphics[width=1.\linewidth]{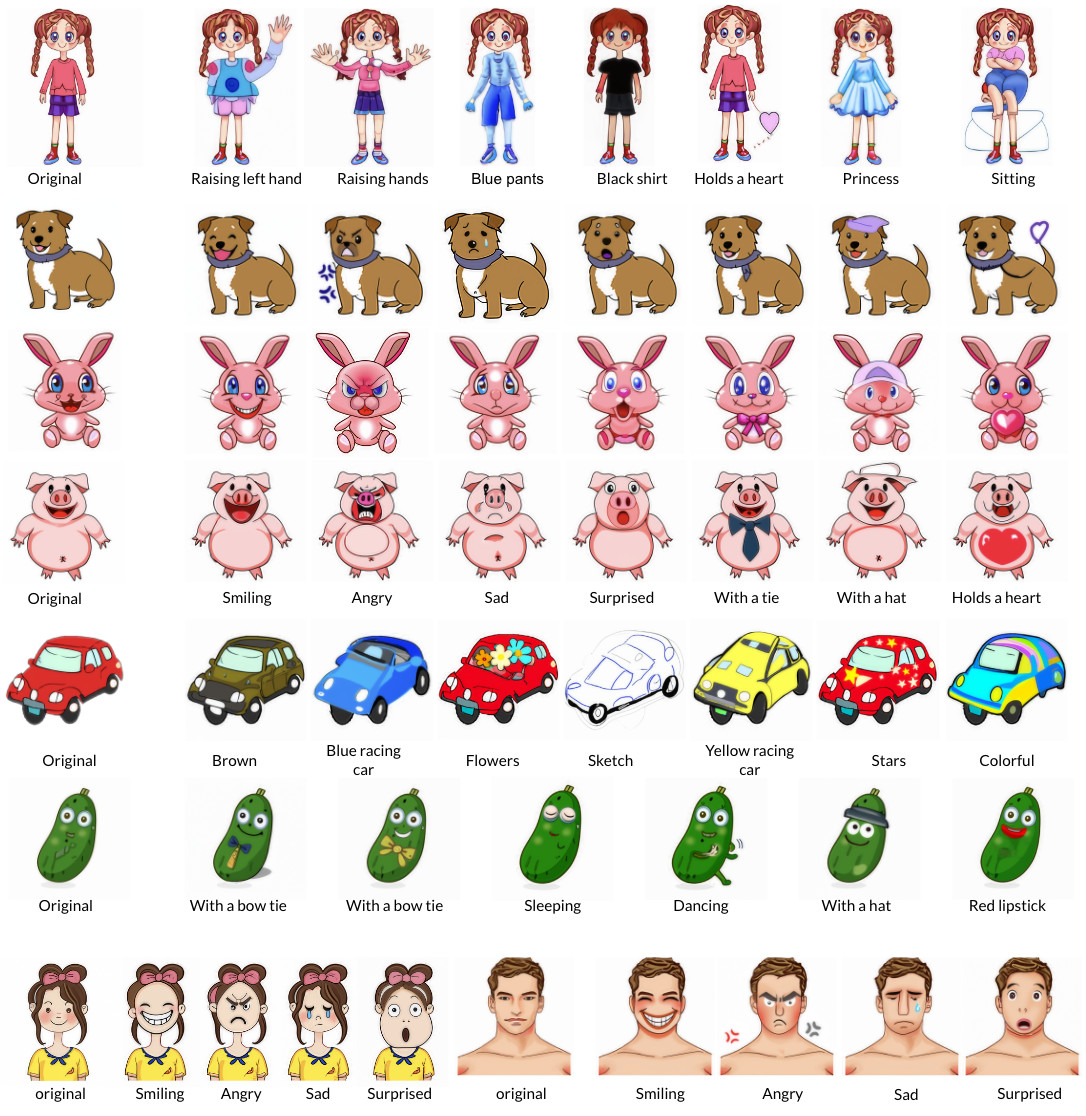} 
\caption{Additional manipulation examples, generated using our model.} 
\label{fig:expression_manipulations}
\end{figure}

\begin{figure}[ht!]
\centering
\includegraphics[width=1.\linewidth]{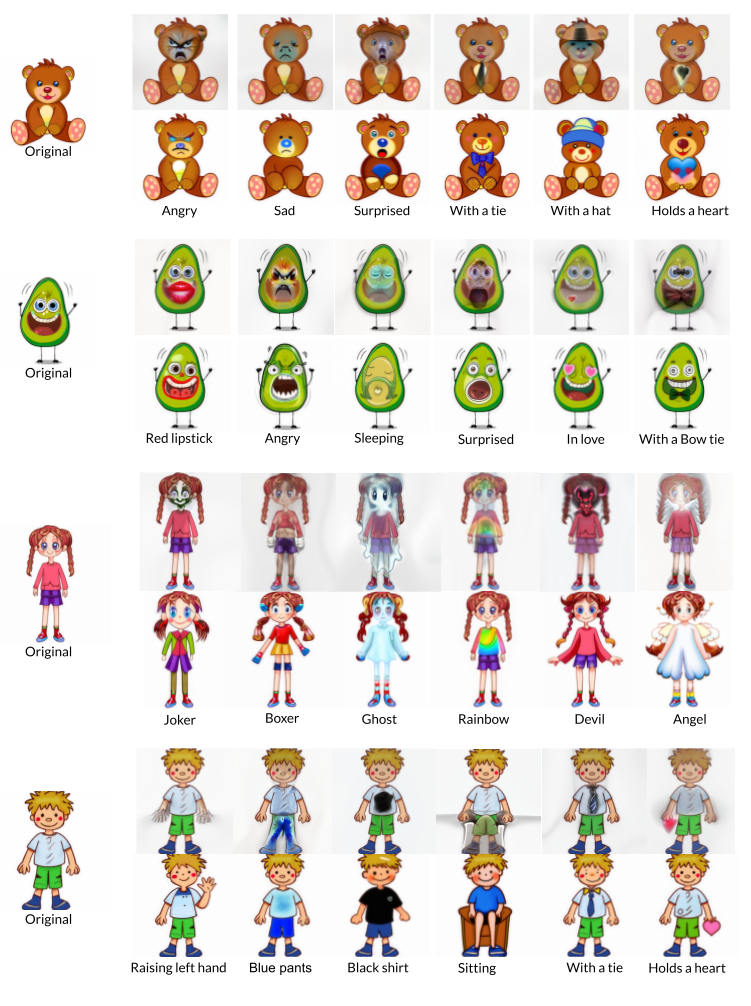}
\caption{comparison to Text2LIVE~\citep{bar2022text2live}. For each input image, the bottom row corresponds to images generated by our model, and the top row corresponds to images generated by the Text2LIVE model.} 
\label{fig:text2live_compare}
\end{figure}

\begin{figure}[ht!]
\centering
\includegraphics[width=1.\linewidth]{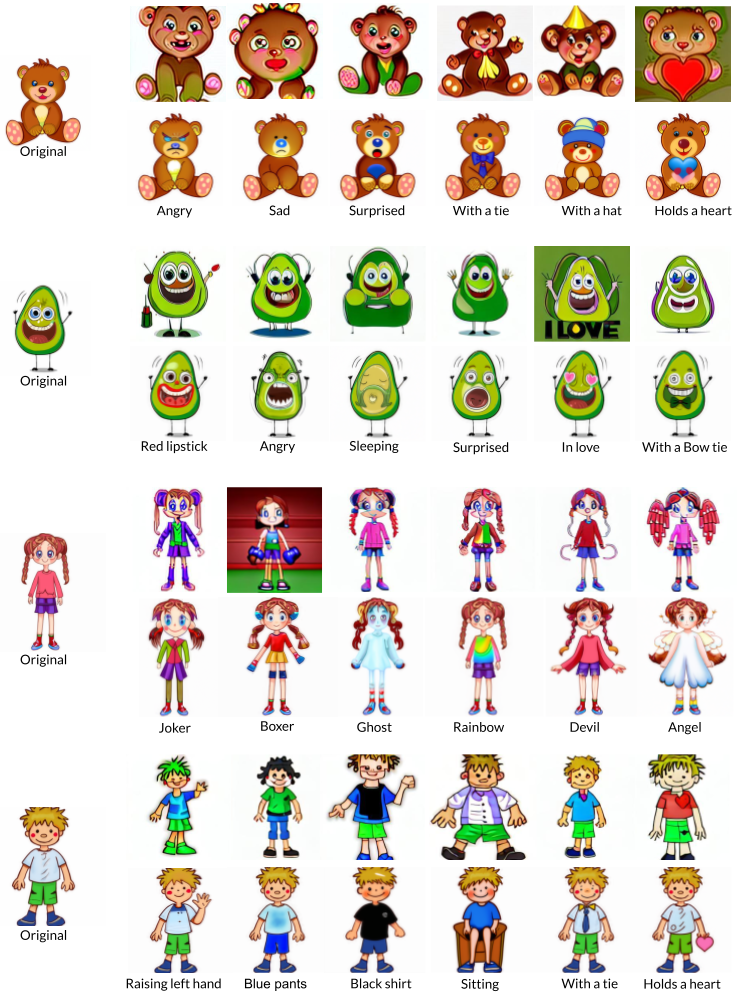} 
\caption{comparison to Textual Inversion~\citep{gal2022image}. For each input image, the bottom row corresponds to images generated by our model, and the top row corresponds to images generated by the Textual Inversion model.} 
\label{fig:textual_compare}
\end{figure}

Additional manipulation examples are provided in Figs.~\ref{fig:expression_manipulations}. The full comparison with the baselines is provided in Fig.~\ref{fig:text2live_compare} and~\ref{fig:textual_compare}. 
We also provide in Fig.~\ref{fig:manipulations_construction} several examples for the process of the manipulated images construction.

\subsection{Additional Samples}
\begin{figure*}[ht!]
\centering
\includegraphics[width=1.0\linewidth]{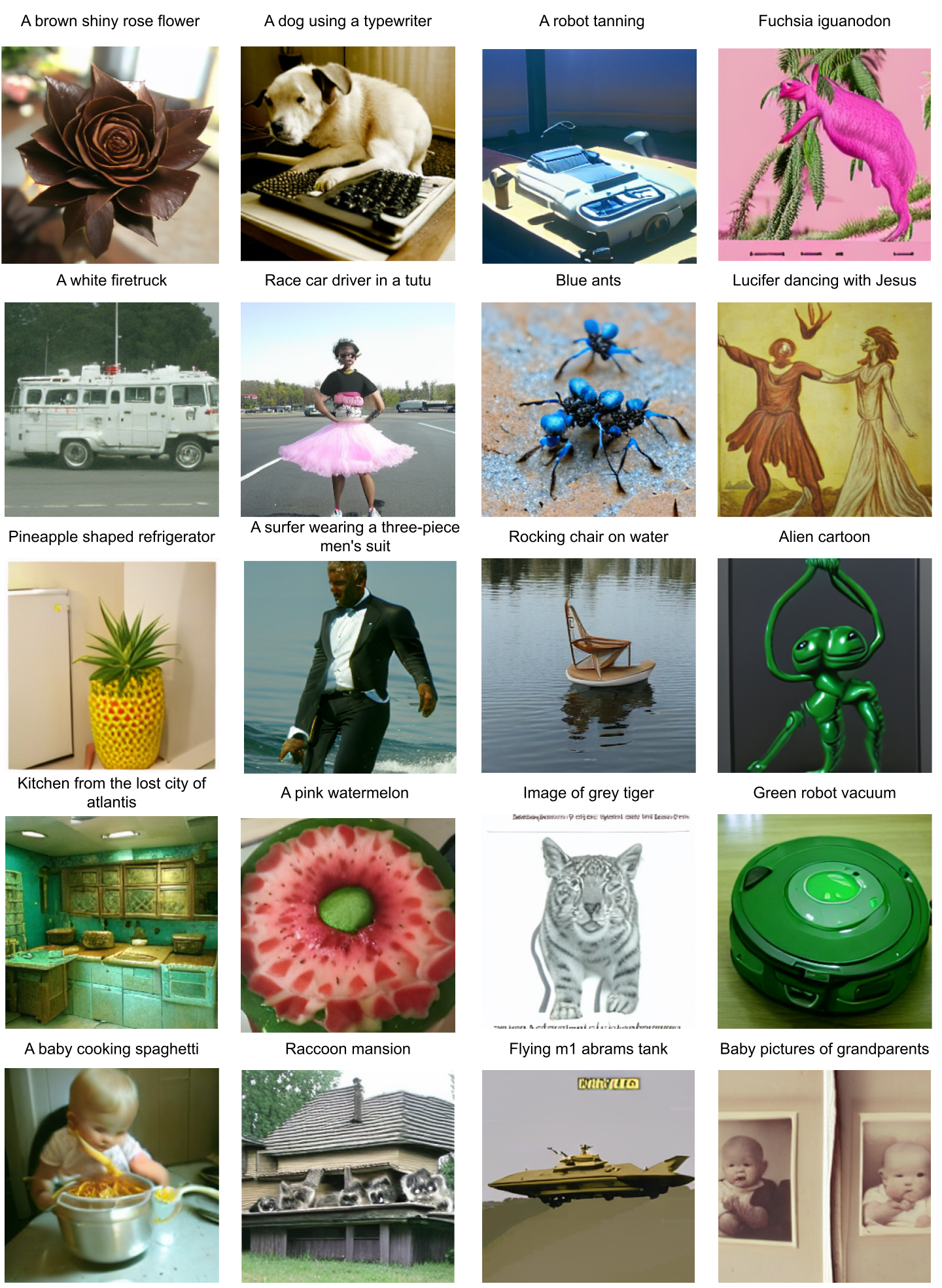}
\caption{Additional samples generated from challenging text inputs using the photo-realistic model} 
\label{fig:cmd_challanging}
\end{figure*}

\begin{figure}[ht!]
\centering
\includegraphics[width=1\linewidth]{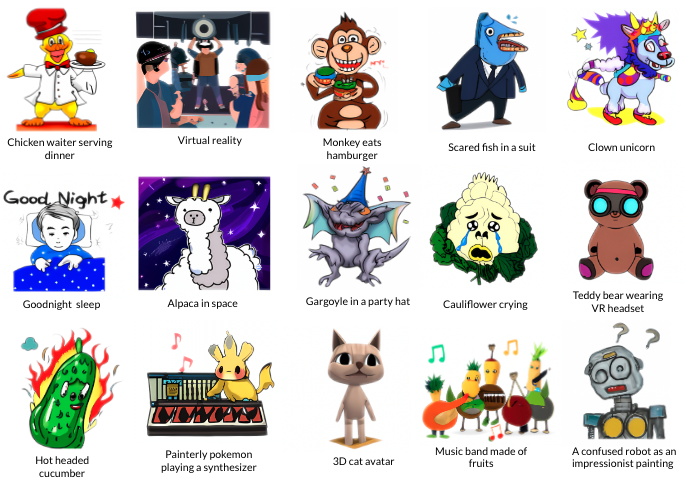}\\
\includegraphics[width=1\linewidth]{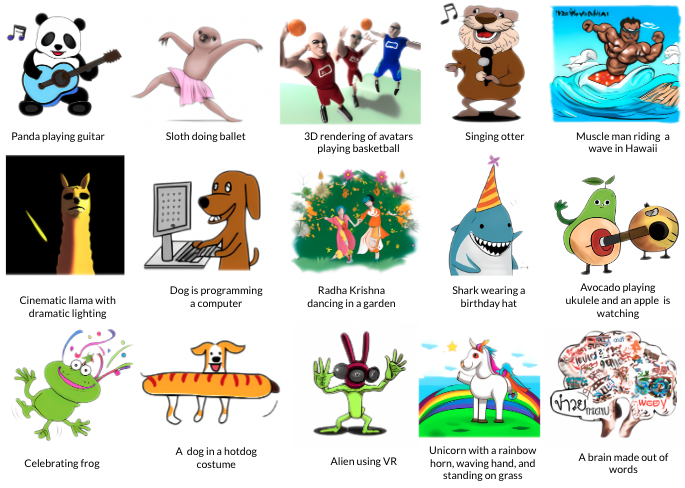}
\caption{A selection of stickers generated using the continuous \knndiffusion model.} 
\label{fig:alpha_stickers_supp}
\end{figure}

\begin{figure*}[ht!]
\centering
\includegraphics[width=0.9\linewidth]{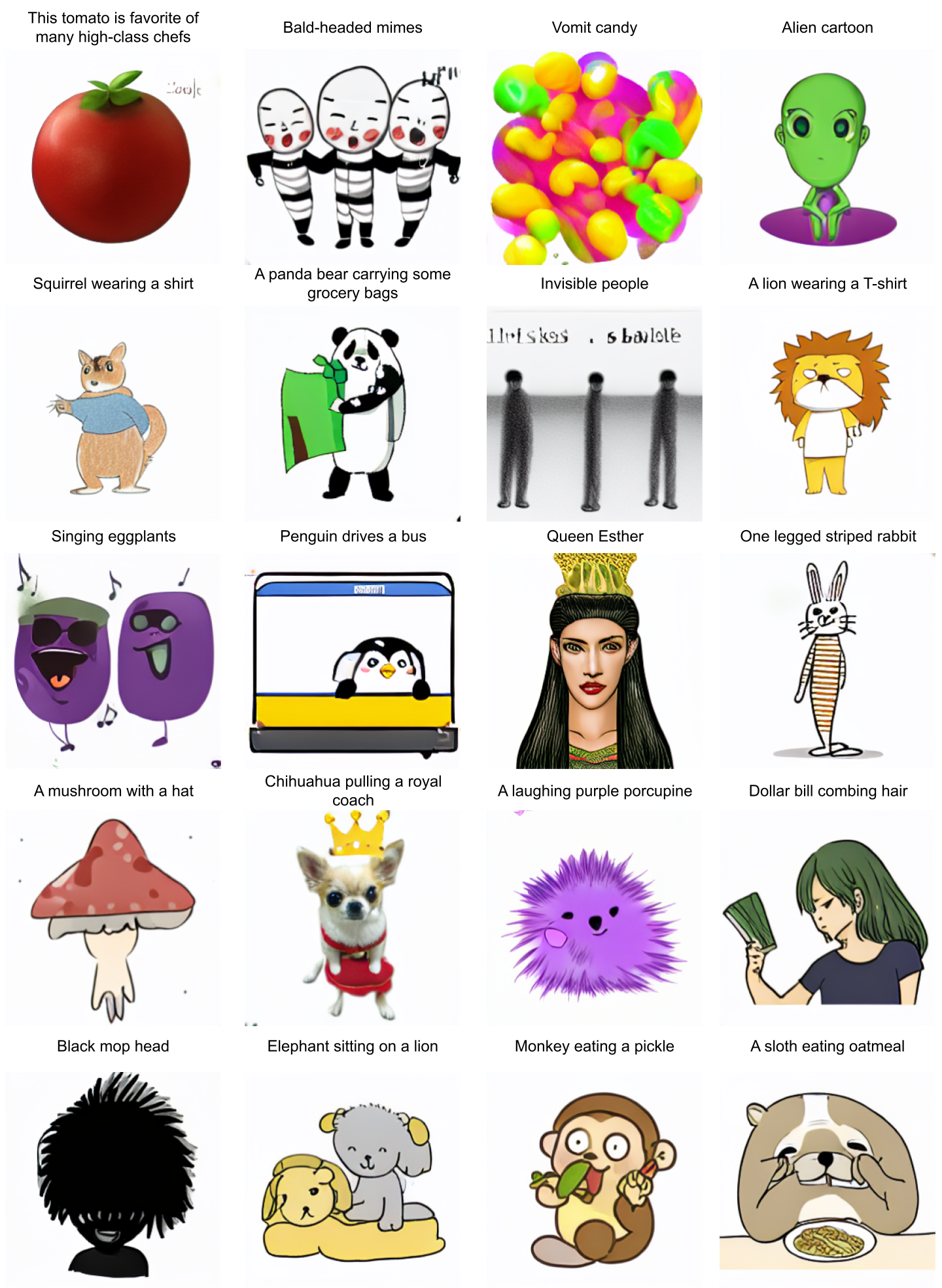}
\caption{A selection of stickers generated using the discrete \knndiffusion model.} 
\label{fig:stickers_challanging}
\end{figure*}

Additional samples generated from challenging text inputs are provided in Figs. \cref{fig:cmd_challanging}, \ref{fig:stickers_challanging} and Fig.~\ref{fig:alpha_stickers_supp}.


\end{document}